\RequirePackage{fix-cm}
\documentclass[smallextended,oneside]{svjour3}       
\smartqed  

\usepackage{lmodern}
\usepackage{anyfontsize}
\usepackage{amsfonts}
\usepackage{aurical}
\usepackage{microtype}
\usepackage{placeins}
\usepackage{graphicx}
\usepackage{multicol}
\usepackage{booktabs} 
\usepackage{array}

\newcounter{turn}
\newcommand{\newturn}{\refstepcounter{turn}(\textbf{\theturn})~\label{ex:\theturn}}
\usepackage{xcolor}
\definecolor{scarlet}{rgb}{1.0, 0.13, 0.0}

\usepackage{etoolbox}
\makeatletter
\patchcmd{\@addmarginpar}{\ifodd\c@page}{\ifodd\c@page\@tempcnta\m@ne}{}{}
\makeatother
\reversemarginpar

\usepackage{mathptmx}      
\usepackage{breakcites}
\usepackage{natbib}

\let\cite\citep

\makeatletter
\newcommand*{\shfttext}[2]{%
  \settowidth{\@tempdima}{#2}%
  \makebox[\@tempdima]{\hspace*{#1}#2}%
}
\makeatother

\usepackage{setspace}
\usepackage{marginnote}

\usepackage{enumitem}
\SetLabelAlign{Center}{\hfil#1\hfil}
\usepackage[framemethod=TikZ]{mdframed}
\usetikzlibrary{shadows}
\usepackage{environ}
\usepackage{varwidth}

\newlength{\MyMdframedWidthTweak}%
\NewEnviron{MyMdframed}[1][]{%
    \setlength{\MyMdframedWidthTweak}{\dimexpr%
        +\mdflength{innerleftmargin}
        +\mdflength{innerrightmargin}
        +\mdflength{leftmargin}
        +\mdflength{rightmargin}
        }%
    \savebox0{%
        \begin{varwidth}{\dimexpr\linewidth-\MyMdframedWidthTweak\relax}%
            \BODY
        \end{varwidth}%
    }%
    \begin{mdframed}[
        backgroundcolor=lightgray, 
        shadow=true, 
        shadowsize=4pt,
        roundcorner=5pt,
        userdefinedwidth=\dimexpr\wd0+\MyMdframedWidthTweak\relax, 
        #1]
        \usebox0
    \end{mdframed}
}

\usepackage{moredefs}
\usepackage{lips}
\usepackage[normalem]{ulem}

\usepackage{xcolor}
\definecolor{perform}{rgb}{0.85, 0.65, 0.13} 
\definecolor{value}{rgb}{0.42, 0.56, 0.14} 
\definecolor{meta}{rgb}{0.0, 0.75, 1.0}    
\definecolor{infer}{rgb}{0.96, 0.29, 0.54} 
\definecolor{gray}{rgb}{0.44, 0.5, 0.56}

\usepackage{newfloat}
\usepackage{placeins}
\usepackage{caption}
\usepackage{footnote}
\usepackage{fnpct}
\usepackage{caption}
\usepackage{wrapfig}
\usepackage{censor}

\usepackage{fancyvrb}

\DeclareFloatingEnvironment[fileext=frm,placement={t},name=Listing]{mylisting}
\usepackage{datatool}
\DTLsettabseparator
\DTLnewdbonloadtrue
\usepackage{underscore}

\usepackage{pf2}
\pfsidenumbers{1}{-30.8em}
\let\pflabel\pflonglabel
\pflongindent

\usepackage{comment}
\let\subparagraph\paragraph
\newcommand\nothing[1]{#1}

\usepackage{marginnote}

\makeatletter
\long\def\@mn@@@marginnote[#1]#2[#3]{%
  \begingroup
    \ifmmode\mn@strut\let\@tempa\mn@vadjust\else
      \if@inlabel\leavevmode\fi
      \ifhmode\mn@strut\let\@tempa\mn@vadjust\else\let\@tempa\mn@vlap\fi
    \fi
    \@tempa{%
      \vbox to\z@{%
        \vss
        \@mn@margintest
        \if@reversemargin\if@tempswa
            \@tempswafalse
          \else
            \@tempswatrue
        \fi\fi
          \rlap{%
            \ifx\@mn@currxpos\relax
              \kern\marginnoterightadjust
              \if@mn@verbose
                \PackageInfo{marginnote}{%
                  xpos not known,\MessageBreak
                  using \string\marginnoterightadjust}%
              \fi
            \else\ifx\@mn@currxpos\@empty
                \kern\marginnoterightadjust
                \if@mn@verbose
                  \PackageInfo{marginnote}{%
                    xpos not known,\MessageBreak
                    using \string\marginnoterightadjust}%
                \fi
              \else
                \if@mn@verbose
                  \PackageInfo{marginnote}{%
                    xpos seems to be \@mn@currxpos,\MessageBreak
                    \string\marginnoterightadjust
                    \space ignored}%
                \fi
                \begingroup
                  \setlength{\@tempdima}{\@mn@currxpos}%
                  \kern-\@tempdima
                  \if@twoside\ifodd\@mn@currpage\relax
                      \kern\oddsidemargin
                    \else
                      \kern\evensidemargin
                    \fi
                  \else
                    \kern\oddsidemargin
                  \fi
                  \kern 1in
                \endgroup
              \fi
            \fi
            \kern\marginnotetextwidth\kern\marginparsep
            \vbox to\z@{\kern\marginnotevadjust\kern #3
              \vbox to\z@{%
                \hsize\marginparwidth
                \linewidth\hsize
                \kern-\parskip
                \marginfont\raggedrightmarginnote\strut\hspace{\z@}%
                \ignorespaces#2\endgraf
                \vss}%
              \vss}%
          }%
      }%
    }%
  \endgroup
}
\makeatother
\let\thesis\nothing
\let\paragraph\nothing
\usepackage{hyperref}
\journalname{Argumentation}
\begin{document}

\title{Argumentation theory for mathematical argument\thanks{Support from the UK Engineering and Physical Sciences Research Council is acknowledged under grants EPSRC EP/K040251/2 (Corneli, Martin and Nesin), EP/J017728/2 (Murray-Rust) and EP/P017320/1 (Pease).}}

\author{Joseph~Corneli \and
        Ursula~Martin \and
        Dave~Murray-Rust \and
        Gabriela~Rino~Nesin \and
        Alison~Pease
}

\authorrunning{~}

\institute{J. Corneli \at
              School of Informatics\\
              University of Edinburgh\\
              \email{joseph.corneli@ed.ac.uk}
           \and
          U.~Martin \at
            Department of Computer Science\\
            University of Oxford
           \and
           D.~Murray-Rust \at
             Edinburgh College of Art\\
             University of Edinburgh
           \and
          G.~Rino Nesin \at
             Computing, Engineering and Mathematics\\
             University of Brighton
          \and
          A.~Pease \at
            Computing, School of Science and Engineering\\
            University of Dundee
}

\date{Received: date / Accepted: date}

\maketitle

\begin{abstract}
To adequately model mathematical arguments the analyst must be able to
represent the mathematical objects under discussion and the
relationships between them, as well as inferences drawn about these
objects and relationships as the discourse unfolds.
We introduce a framework with these properties, which has been 
used to analyse mathematical dialogues and expository texts.
The framework can recover salient elements of discourse at, and within, the sentence
level, as well as the way mathematical content connects to form larger
argumentative structures.  We show how the framework might be used to
support computational reasoning, and argue that it provides a more
natural way to examine the process of proving theorems than do
Lamport's structured proofs.
\keywords{inference anchoring theory \and mathematical practice \and mathematical argument \and structured proof}
\end{abstract}

\clearpage

\section{Introduction}\label{intro}

The representation of mathematical knowledge and inference in appropriate formal logical frameworks is well-understood and the subject of much research. Computational tools to support this through proof checking, automatic theorem proving, and computer algebra are well-established, though they require  formal, computationally explicit, content as input. However, the existing mathematical literature, particularly informal mathematical dialogues, and expository texts, is opaque to such systems, which  cannot currently handle  the variety of activities typically involved in producing such knowledge and proofs, such as, for example, exposition and argument that concerns making conjectures, forming concepts, and discussing  examples and counterexamples.  Our goal is to bridge this gap through devising an expressive modelling language that is closely related to the way mathematics is actually done.

Our approach to modelling such content is inspired by the general-purpose argument modelling formalism Inference Anchoring Theory (IAT), introduced by \citet{reed2011dialogues}. As its name suggests, IAT anchors logical inferences in discourse. IAT has been applied to mediation 
\cite{janier2017towards}, debates \cite{budzynska2014model}, and to paradoxes in ethotic argumentation \cite{budzynska2013circularity}, along with other real-world dialogues \cite{budzynska2013towards}. The \emph{Inference Anchoring Theory + Content} (IATC) framework we introduce is based on IAT, but with several significant modifications.  Most fundamentally, IATC is designed to bring to the surface the structural features inherent in mathematical content.

IATC could be overlaid upon formally specified contents, where these are available.  Lamport's ``Temporal Logic of Actions+'' (TLA+) \cite{lamport1999specifying,lamport2014tla2} is one such formalism that could be used to model content-level expressions.  Higher-level discourse structure would then be exhibited somewhat along the lines of Lamport's own semi-formal ``structured proofs'' 
\cite{lamport1995write,lamport2012write21st}.  However, unlike structured proof, IATC does not aim to reshape the way people do mathematics, but to model it more exactly.  As such, it constitutes groundwork for a future generation of computer systems that can collaborate with mathematicians and students in a way these potential users already understand. \citet{epstein2015wanted} highlights the ``extent to which a person believes that her work experience or product has been facilitated or improved by the collaboration'' as a key evaluation metric for assessing collaborative intelligent computer systems. The key metric at this stage is more basic, namely, we are interested in the degree to which IATC can represent real-world examples of mathematical practice in a way that can make them accessible to computational reasoning. After introducing the modelling approach, we use several examples to show that IATC is indeed satisfactory in this regard.

\begin{itemize}
\item Our first example is a school-level challenge problem that was
  presented in a public lecture by the mathematician Timothy Gowers
  \cite{gowers-talk}.  The lecture aimed to motivate and contextualise
  a project, then beginning, to develop mathematical software that
  ``operate[s] in a way that closely mirrors the way human
  mathematicians operate'' \cite[p.~255]{ganesalingam2016fully}.  The
  reasoning needed to solve the challenge problem remains beyond the
  scope of the computational method that Ganesalingam and Gowers
  ultimately published, but it is both sufficiently simple and
  sufficiently realistic to introduce the practical aspects of
  working with IATC.
\item Our second example is a question posed on the online Q\&A forum
  MathOverflow, together with the ensuing dialogue.  MathOverflow is
  part of the Stack Exchange network of community question-and-answer
  websites, which is particularly popular with software developers.
  The MathOverflow sub-site is devoted to discussions about
  research-level questions in mathematics.  Such discussions are very
  different from the textbook-style proofs treated by
  \cite{ganesalingam2016fully}, and we discuss the considerations that
  such discussions would impose on computational modelling efforts.
\item \paragraph{MiniPolymath 1 through 4 were part of a series of
  experiments in collaborative online mathematics known as ``Polymath
  projects'' \cite{polymath-wiki}.}  While other projects in the series
  tackled novel research, the problems in the MiniPolymath subseries
  were drawn from the Mathematical Olympiad, a premier competition for
  pre-college students.
Six problems are given, and the examination takes place over
two days with three problems to be solved each day.
Whereas individual Olympiad participants
frequently fail to solve three challenge problems in the
four-and-a-half hours allotted for that purpose, all four of the
collaborative MiniPolymath efforts generated a solution.
However, it should be noted that some of these solutions
took more than 24 hours to develop.
IATC can help us understand how the proof efforts progressed, and can
potentially help us understand why they were (mathematically)
successful.
\end{itemize}

The plan of the work is as follows.
\textsection\ref{background} reviews previous research on mathematical
argument, presents a brief introduction to Inference Anchoring Theory, and
describes Lamport's structured
proofs as an example of the state of the art for
modelling informal mathematical knowledge.
\textsection\ref{iatc} introduces IATC, describes the grammar of IATC markup,
and describes the differences between this language and IAT.
\textsection\ref{iatc-examples} presents our analysis of the examples
outlined above, which have been marked-up with IATC in order to
illustrate the relevant modelling concerns.
\textsection\ref{discussion} summarises and reviews the contribution,
situates our work in relationship to the broader literature, and
outlines potential directions for further work.

\section{Background}\label{background}

\thesis{In this section we state what we
mean by argumentation, and survey previous
  research on argumentation in mathematics
  (\textsection\ref{argmath}).}  We then describe Inference
Anchoring Theory (\textsection\ref{iat}) and structured proof
(\textsection\ref{sec:structured-proofs}), two landmarks that
guide our effort.

\subsection{Argumentation and mathematical arguments}\label{argmath}

Our approach to argument builds on Buzynska and Reed's
Inference Anchoring Theory (IAT), which we describe in
Section \ref{iat}.  The specific conception of \emph{argument} that
underlies IAT is as follows:
\begin{quote}
[A]rguing can be interpreted as an illocutionary act that comes about as
the result of a relation between uttering a premise and uttering a
conclusion, thus mirroring the logical structure of inference[.]  \cite[p.~146]{Reed2017}
\end{quote}
\citet{reed2011dialogues} note that in everyday language the term
``argument'' is used to describe a particular kind of interaction
as well as the shared understanding extracted from these
interactions, as ``evidence'' or ``proof.''
The purpose of IAT is to make the links between discourse and reasoning explicit.

Concerning argumentation in a mathematical context, \citet[p.~39]{pedemonte2007} argues that
``analysis of the `content' is not sufficient to analyse all the cognitive aspects in
the relationship between argumentation and proof.''
A large part of mathematical discussion is in essence meta-discussion
about meta-level objects, such as proof strategies that are
suggested on the fly and debates about whether these strategies are
likely to work as intended.

\citet{mercier2011humans} distinguish
arguments from inferences: 
only in the case of arguments ``the reasons for drawing this
conclusion on the basis of the premises are (at least partially)
spelled out'' (p.~58).   
By contrast, formal mathematics is typically based on the reductive assumption that
``mathematical reasoning may be identified with classical, deductive
inference'' \cite[p.~25]{aliseda2003mathematical}.
However, everyday mathematical reasoning plainly involves more than just proof steps.
Here are two examples of familiar patterns of reasoning that appear in MiniPolymath 3:
\begin{description}
\item[\emph{argument from authority}]
``My bachelor thesis supervisor said that one can't use the word cardinal if we talk about finite sets. One has to use the words `number of elements'{''}
\cite[\href{https://wp.me/pAG2F-41\#comment-3418}{19 July, 9:46 pm}]{tao2011imo}.
\item[\emph{argument from analogy}]
``Let me check that I got the example correctly: is this `a point inside a regular polygon'? Isn't it established in an early comment that the example of a point inside an equilateral triangle indeed visits \emph{all} the points? Can you clarify the difference here?''
\cite[\href{https://wp.me/pAG2F-41\#comment-3378}{19 July, 9:19 pm}]{tao2011imo}.
\end{description}
The word ``argument'' has been attached to several distinct kinds of
mathematical artefacts and activities.  This term may indicate
\emph{proofs} \citet{gasteren1990shape},
\emph{informally-presented proofs} \cite{tanswell2015problem},
\emph{proof sketches} \cite{lamport1995write},
\emph{aspects of reasoning that are not addressed by formal deduction} \cite{Aberdein2013} and \emph{elements of persuasive discussion} \cite{Zack2001}.

Some theorists have expressly contended that proofs are not arguments:
this is because proofs offer certainty, while arguments cannot
\cite{Dufour2013}.
Nevertheless, communication of reasons and reasoning
can be found throughout mathematical practice.
\citet{pedemonte2007} highlights the use of inductive and abductive logic
as well as deduction in mathematical processes that move from
``from conjecturing to the construction of proof [to] the proof as
product,'' and in which ``content rather than formal criteria''
can guide the proving process.  \citet{Dufour2013} gives examples
of argumentation ``not only before and during the proof but also after,
at least as long as it can be criticized'' (p.~74).
Other scholars have observed features such as these:
\begin{itemize}
\item Published mathematical writing tends to be particularly explicit about
reasons and conclusions \cite[p.~149]{dove2009towards}.
\item 
Not only the \emph{Prover} but also the \emph{Skeptic}
``has an important role to play, namely to ensure that the proof is persuasive, perspicuous, and valid'' \cite[p.~2618]{novaes2016reductio}.
\item On the way to a proof, degrees of
  confidence about the conclusions to be drawn may be discussed \cite[p.~17]{inglis2007modelling}.
\item Mathematical meanings need to be interpreted, and this tends to be a struggle  \cite[p.~360]{vanOers}.
\end{itemize}

\paragraph{\citet{Carrascal2015} provides an excellent survey of recent thinking about argument in mathematics, highlighting its connections with mathematical practice.}
Carrascal advises: ``in
order to learn more about the nature of mathematical practice and how
its products are evaluated, we should be looking at real examples of
this practice.''  She points to \citet{pease-and-martin} as a notable
example in this genre.  Once we have developed a suitable apparatus,
Section \ref{iatc-examples} will tackle several
real-world examples, including a detailed reexamination of the
dataset studied by Pease and Martin.

``Blog maths'' \cite{barany2010b} and other online discussions,
for example, on the
question-and-answer site MathOverflow, can ``tell us
about mathematicians' attitudes to working together in public'' as
well as the ``kinds of activities that go on in developing a proof''
\cite{Martin2015}.
In the process of creating a proof or mathematical theory, divergent
understandings are negotiated using shared concepts, definitions, and
standards for proof, even as the concepts evolve.  Along these lines,
\citet{lak} used the methods of structured and abstract argumentation
to formalise the theory of informal mathematics developed in
Lakatos's \emph{Proofs and Refutations}
\citeyearpar{lakatos2015proofs} as a set of rules for turn-taking in a
dialogue game.  This work shows that formally specified and fully
implemented argumentation tools can be brought together and applied to
a specific, demanding, domain of human reasoning.\footnote{\label{fn:technical-details}The
  dialogue game defines ordered operations on a shared information state
  represented in the Argument Interchange Format (AIF)
  \cite{lawrence2012aifdb}, which is then interpreted by The Online
  Argument Structures Tool (TOAST) \cite{Snaith2012} and passed on to
  DungOMatic \cite{snaith2010pipelining} to calculate the grounded
  extension, which in this case represents the currently accepted,
  collaboratively constructed, proof or theory under discussion.}
\citet{dauphin2018aspicend} produced a similar model of
natural-deduction style arguments, explanations,
and the ``\emph{prima facie} laws of logic'' such as may be debated in
work on mathematical foundations.
These prior efforts focus on developing rules that give a plausible codification
of mathematical process.  Our concern is different,
but complementary.  We are interested in a better
understanding of what is actually said in mathematical arguments, and on
the reasoning that is conveyed.
Accordingly, we will adapt a general-purpose argument modelling
approach, \emph{Inference Anchoring Theory}, which is described in the
following section.

\subsection{Inference Anchoring Theory} \label{iat}

\thesis{Inference Anchoring Theory (IAT) is used to model the logical
  relationships between the propositional contents of 
  utterances made in dialogues \cite{budzynska2011speech}.}  As noted
by \citet{Reed2017}, the inspiration for developing IAT lies in
earlier work on representing dialogue in the Argumentation Interchange
Format.

\paragraph{IAT is grounded in a notion of \emph{dialogical relations} that formalise the informal ``conventions and norms that dictate the flow of dialogue'' \cite{snaith2016dialogue}.}
Per \citet{budzynska2011speech}, these dialogical relations are also
referred to as ``\emph{transitions},'' a term that is meant to recall
the notion of transitions between operating states in a finite state
machine.  Indeed, when the norms have been fully codified in a
\emph{dialogue protocol}, the transitions are exactly described by a
finite state machine.\footnote{In such a setting the formal
  argumentation-theoretic techniques and tools mentioned in Footnote
  \ref {fn:technical-details} can be applied, though IAT models are
  not required to be fully formal in this regard.}  Content
relationships are typically identified by matching locutions against
known argument schemes, e.g., an `Argument from Positive Consequences'
is associated with two transitions, `challenging' and `substantiating'
\cite{walton08}.
\citet{budzynska2014towards} describe Inference Anchoring Theory in terms of three components:
\begin{quote}
\begin{enumerate}[align=Center,label=(\roman*),labelsep=4pt,leftmargin=*,labelwidth=!]
\item\label{transitions-item} relations between locutions in a dialogue, called \emph{transitions};
\item\label{contents-item} relations between sentences (propositional \emph{contents} of
  locutions); and
\item\label{connections-item} \emph{illocutionary connections} that link locutions with their
  contents.
\end{enumerate}
\hfill \cite{budzynska2014towards}, emphasis added
\end{quote}

\begin{figure}[h]
\centering
\begin{minipage}{.4\linewidth}
\centering
\includegraphics[width=\linewidth,trim=9cm 7cm 8.5cm 6.9cm, clip]{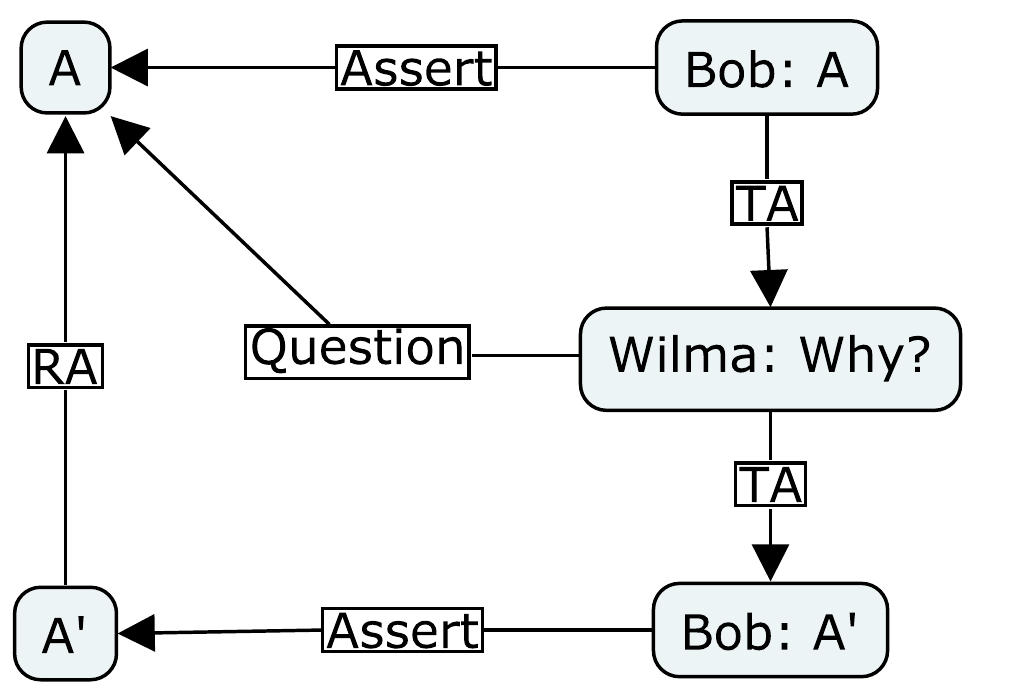}
\caption{IAT diagram for the conversation `$A$'/`\emph{Why?}'/`$A^\prime$'.\label{fig:TypicalIAT}}
\end{minipage}
\hspace{.1\linewidth}
\begin{minipage}{.4\linewidth}
\centering

\includegraphics[width=\linewidth,trim=9cm 7cm 8.5cm 6.9cm, clip]{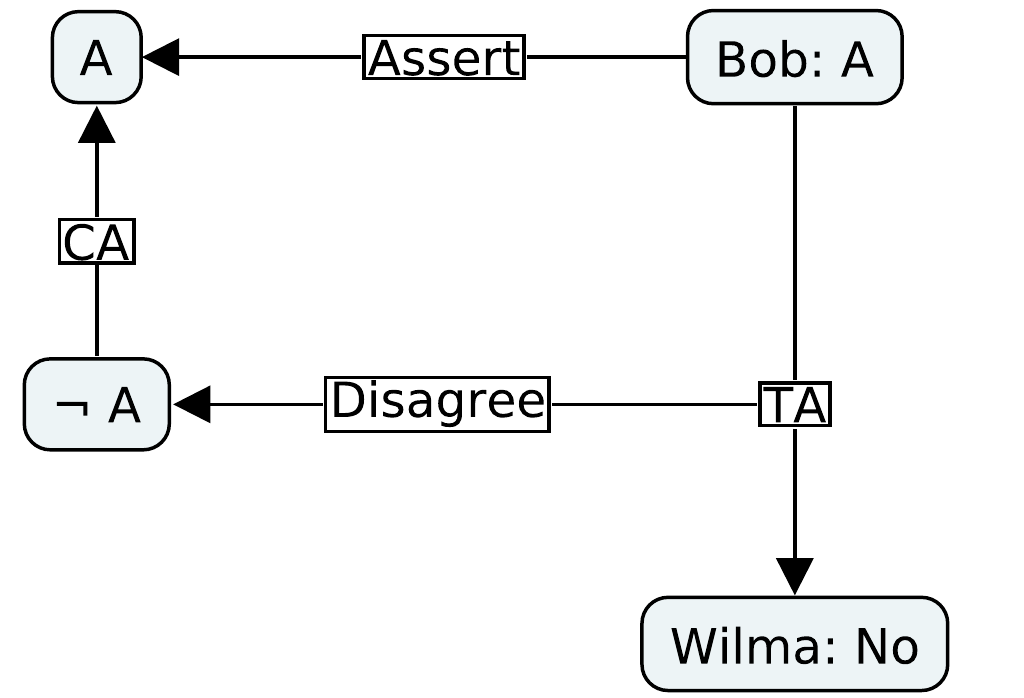}
\caption{IAT diagram for the conversation `$A$'/`\emph{No}'.\label{fig:A-not-A}}
\end{minipage}
\end{figure}

In Figures \ref{fig:TypicalIAT} and \ref{fig:A-not-A}, below, ``TA''
stands for a \emph{default transition}, ``RA'' stands for
\emph{application of rule of inference}, and ``CA'' stands for
\emph{default conflict}.  That is to say, there is no explicit formal dialogue
protocol attached to these two examples.

Figure \ref{fig:TypicalIAT} is a typical example of an IAT analysis.
Figure \ref{fig:A-not-A} illustrates a feature
that was not directly mentioned in the list \ref{transitions-item}--\ref{connections-item}, above;
specifically,
this figure uses an \emph{{`}implicit{'} speech-act}
to anchor propositional content on a transition rather than a locution.
Here, when a speaker asserts `$A$' and their interlocutor says
`\emph{No}', the logical content `$\neg A$' is
attached to the transition, rather than to the negating word.
The basic rationale is that the locution `\emph{No}' cannot be made sense
of without the preceding context.  There has been some debate about what to do about this.
\citet{botting2015inferences} says that the choice to
anchor arguments on transitions is a conceptual mistake.  
However, for the creators of IAT, the reason illocutionary acts can
be rooted on dialogical relations follows
\begin{quote}
\ldots directly from pragma-dialectical analysis which views
the speech act of assertion [\lips] as occurring at the `sentence'
level, and the speech act of argumentation as occurring at a `higher
textual level.' \cite{budzynska2011speech}
\end{quote}

\citet{visser2011correspondence} describe the theoretical
considerations in more detail.
The pattern common to both Figure \ref{fig:TypicalIAT} and Figure
\ref{fig:A-not-A} is that allowable inferences are governed by
dialogue norms.  In Figure \ref{fig:TypicalIAT}, for instance, we
would not immediately know that `$A^\prime$' is intended to \emph{support} `$A$'
without Wilma's intermediate question which explicitly requested such
support.  Given the context, the intended
inference is is clear.
Thus, both examples serve to illustrate that
\begin{quote}
the connection between locutions in a dialogue has an inferential
component beyond any that may hold between the contents of those
locutions \cite{reed2011dialogues}.
\end{quote}
In short, IAT studies ``the way in which the rules of dialogue
influence the construction of argument''
\cite{budzynska2016theoretical}.

Although the specific example in Figure \ref{fig:A-not-A} is very
simple, the following general observation on dialogue norms is useful for thinking about how the
conversation might continue from the point it has reached so far:
\begin{quote}
[T]here is an asymmetry between the \emph{production} of arguments,
which involves an intrinsic bias in favor of the opinions or decisions
of the arguer whether they are sound or not, and the \emph{evaluation}
of arguments, which aims at distinguishing good arguments from bad
ones. \cite[p.~72]{mercier2011humans}
\end{quote}
If the conversation were to continue, Wilma would typically have the
burden of justifying her rejection of `$A$', which might be done with
counterarguments that would dig into the details of `$A$' looking for
flaws (\emph{ibid.}, p.~67); in addition, she might begin to make a
case for an alternative position, `$B$'.  These considerations point to
the direction we will be taking with IATC.

\paragraph{Our main strategy will be to supplement IAT with an explicit register for content.}
Alongside \ref{transitions-item}--\ref{connections-item}, above, we
introduce:
\begin{quote}
\begin{itemize}[label=(\roman*)]
\item[(iv)] a model of non-propositional content, namely of the
  mathematical objects under discussion, and the relations between
  them.
\end{itemize}
\end{quote}
We will describe the implications of this addition in detail in
Section \ref{iatc}, along with some other adaptations to IAT that we
have found useful in mathematical settings.  One of the implications
is that in the current work we do not need to emphasise transitions---of
either the explicit or implicit variety---since a more explicit
treatment of content gives us another way to manage context
relationships.
\subsection{Lamport's structured proofs} \label{sec:structured-proofs}

\thesis{\emph{Structured proofs}, as described by
  \citet{lamport1995write,lamport2012write21st}, inhabit the middle
  ground between formal and informal mathematics, and
provide a useful point of reference for our work on IATC.}
Structured proofs offer a notational strategy that is a ``refinement
of [\lips] natural deduction'' \cite{lamport1995write}.  While the
proofs represented using this system are not required to be strictly
formal, the language of structured proofs has evolved together with
Lamport's work on a formal language and corresponding proof checking
system, the ``Temporal Logic of Actions+'' ({TLA}$^{+}$), which is used
to model concurrent systems
\cite{lamport1999specifying,lamport2014tla2}.\footnote{Only a few
  of the keywords available in the latest version of
  TLA\(^{+}\) appear in the structured proof notation.  Per
  \citet{lamport2015tlahyperbook}, the full list of TLA\(^{+}\)
  keywords is as follows.  Those which are also used in structured
  proofs are decorated with underlining: \uline{\texttt{assume}}
  \ldots\ \uline{\texttt{prove}} \ldots, \texttt{boolean},
  \texttt{by}, \uline{\texttt{case}}, \texttt{choose},
  \texttt{constant} (synonymously, \texttt{constants}),
  \texttt{corollary}, \texttt{def}, \texttt{define}, \texttt{domain},
  \texttt{else}, \texttt{except}, \texttt{extends}, \texttt{have},
  \texttt{hide}, \texttt{if}, \texttt{instance}, \texttt{lambda},
  \texttt{lemma}, \uline{\texttt{let}} \ldots\ \uline{\texttt{in}}
  \ldots, \uline{\texttt{new}}, \texttt{omitted},
  \uline{\texttt{pick}}, \texttt{proposition}, \texttt{recursive},
  \texttt{subset}, \uline{\texttt{suffices}}, \texttt{take},
  \texttt{theorem}, \texttt{unchanged}, \texttt{union}, \texttt{use},
  \texttt{variable}, \texttt{witness}.}  Structured proofs are,
specifically, structured as a strict hierarchy of lemmas.
An example appears later on in this paper, in Figure \ref{fig:gowers2012-ala-lamport},
which we will use to illustrate the similarities and differences with
IATC.

For now, we comment that while the use of strict hierarchies
is not representative of the way proofs are usually
constructed in day-to-day practice, Lamport has proposed that
structured proofs can assist in proof development, e.g., by helping to
bring errors to the surface.  However, they do not necessarily make the job of
the reader easier: Lamport \citeyearpar[p.~20]{lamport2012write21st} quotes a
referee who had read one of his structured proofs:
\begin{quote}
The proofs [\lips] are lengthy, and are presented in a style which I
find very tedious.  [\lips] My feeling is that informal proof sketches
[\lips] to explain the crucial ideas in each result would be more
appropriate.
\end{quote}
Unlike structured proofs, IATC is intended to express the typical processes
by which proofs are generated in standard practice, rather than
make the process of proving and reading proofs easier.
It would nevertheless be compatible with our
aims to include formal statements in {TLA}$^{+}$ (or some other
language) in IATC's content layer.

\section{Inference Anchoring Theory + Content}\label{iatc}

\thesis{IATC has many things in common with IAT, but should not be
  seen as a strict addition to the earlier theory.}  Adding explicit
models of content and discussions about content prompts several adaptations.  In this section
we describe these adaptations, and introduce the IATC
modelling language.

\paragraph{Several important requirements arise from the features of the mathematics domain.}
As we saw above, IAT is concerned with anchoring propositions to
utterances and with mapping the logical relationships that obtain
between them.  However, various mathematical objects---\citet{larvor2012think} mentions ``diagrams, notational expressions,
physical models, mental models and computer models''---are more
comfortably thought of as non-propositional in nature.  
Discussions about proofs have been theorised
formally using the notion of \emph{proof plans}, which are constructed
and transformed using explicit heuristics and tactics
\cite{bundy1988use}.  However,
\citet[pp.~63-64]{fiedler2007argumentation} have suggested that
existing work with proof plans cannot be straightforwardly adapted
from machine-oriented to human-oriented contexts, because proof plans
are, from a potential human reader's perspective, overly detailed,
with insufficient structural abstraction.  By contrast, a language like
IATC is charged with expressing ``strategic arguments that are
meaningful to humans'' \cite[p.~68]{fiedler2007argumentation}.
Nevertheless, as important as strategic reasoning is, low-level
mathematical content seems to be even more fundamental.

\paragraph{We see the first-class role that content plays in mathematical discourse when new terms are introduced and referred to, for example.}
Thus, the editor's introduction to
\citet{karttunen1976discourse} notes the following:
\begin{quote}
\ldots informal notational practise of mathematicians, who will write
an existentially quantified formula (say, $(\exists e)(\forall x)(xe =
ex = x)$, as one of a set of postulates for group theory) and
thenceforth use the variable bound by the existential quantifier as if
it were a constant as when they will write the next postulate ($\forall
x)(\exists x^{-1})(xx^{-1} = x^{-1}x = e)$. [punctuation modified]
\end{quote}
Karttunen's concept of ``discourse referents,'' illustrated in the
quote above, underlies Discourse Representation Theory \cite{kamp1993discourse}
and its extensions.
While the developers of IAT acknowledge the generality of Structured
Discourse Representation Theory (SDRT), in particular, they criticise
it for making ``assumptions of context-independent semantics''
\cite{budzynska2016theoretical}.
Nevertheless, DRT has been successfully applied to model some aspects
of mathematical discourse,
and we will discuss that work further in Section \ref{discussion},
and contrast it with our orientation here.

For now, we emphasise that IATC differs from IAT in its approach to
context.
Specifically, IATC sets the notion of dialogical relations to one side, and
instead connects locutions to each other directly in the content and
intermediate (meta-discussion) layers. 

\begin{wrapfigure}{l}{.49\textwidth}
\vspace{-1.05cm}
\hspace{-.2cm}\includegraphics[width=.5\textwidth,trim=5cm 6cm 5cm 5.5cm, clip]{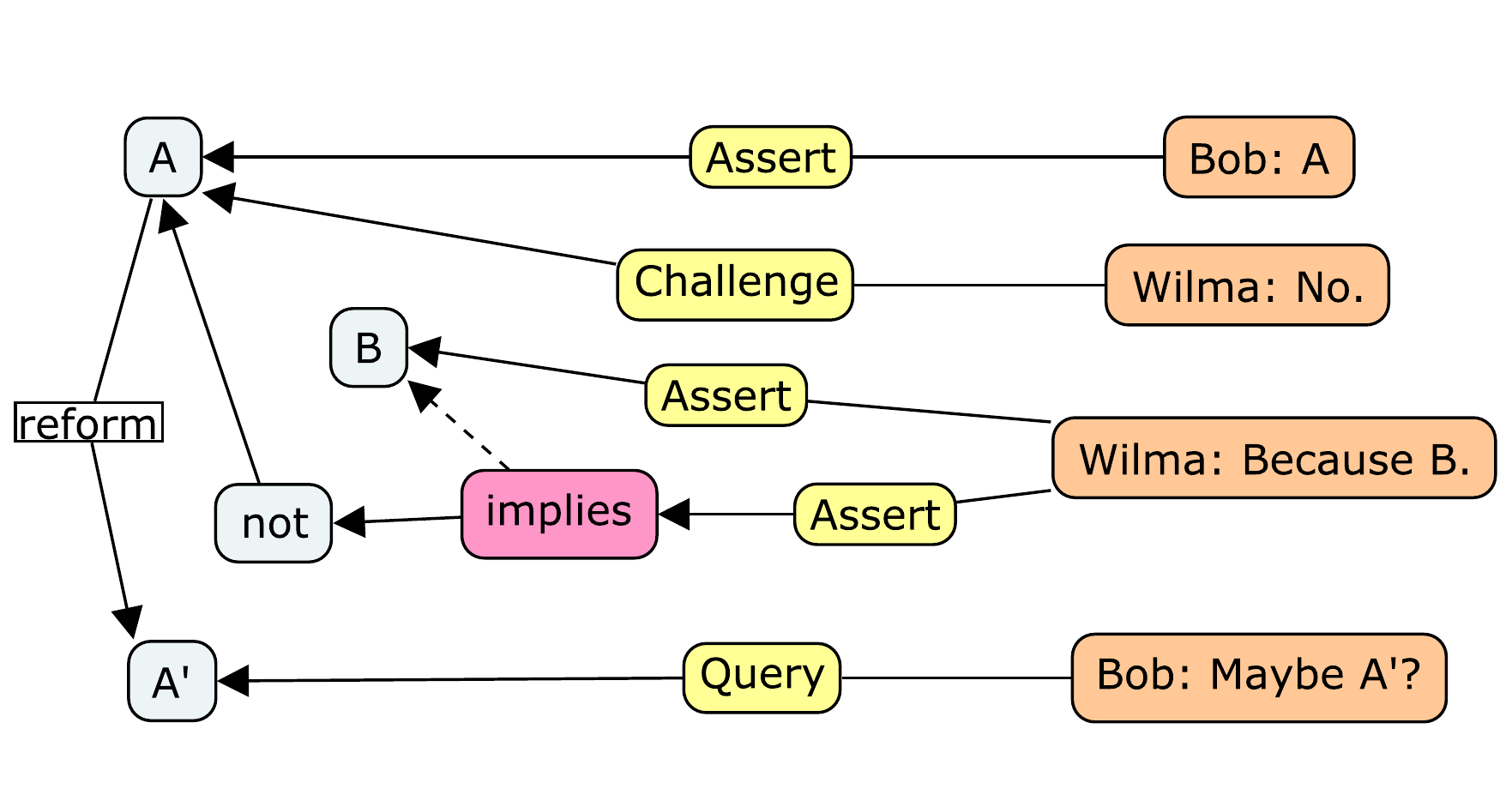}

\caption{Simple IATC diagram exhibiting an assertion, a refutation, a counterexample, and a reformation.\label{fig:A-Counterexample}}
\vspace{-.6cm}
\end{wrapfigure}

Before we describe the language
in detail, we present a simple example, Figure \ref{fig:A-Counterexample}, which reanalyses and extends the `$A$'/`\emph{No}' dialogue from Figure \ref{fig:A-not-A}.  The first two
dialogue moves in these two examples are identical.

Here, rather than connecting `\emph{No}' to a `$A$' with a transition, we connect
it directly to the previously modelled content, $A$, via a
`\texttt{Challenge}' illocution.  From there, we continue
to use the content and intermediate layers to explicitly model
interconnections.  For example,
`$B$' does not simply conflict with $A$, but rather presents a
warrant for ``not $A$'', modelled here using the two-parameter
`\texttt{implies}' relation.

With these changes in place, dialogue relations could in principle
be reintroduced.
For example, `\emph{Because $B$}' could be seen
to `substantiate' the previous utterance, `\emph{No}', as a communicated
reason for rejecting $A$.
Nevertheless, in the current work we continue to leave these links out, on the
basis that we do not yet have a detailed theory of the norms of mathematical dialogue.
The Lakatosian model developed by \citet{lak}, for example, only covers
a limited subset of the rules
and norms involved, specifically, those dealing with conjectures,
lemmas, and the production and evaluation of counterexamples.
By interconnecting contents in the content layer
and through intermediate relations,
we are able to make an explicit model of the logical structure of mathematical arguments.  Such models could potentially inform a subsequent analysis of the associated dialogue structures.

For example, the long-range \texttt{reform} connection from $A$ to $A^\prime$ in
our content analysis would suggest a corresponding long-range transition from
Bob's first to his last statement in the dialogue.
However, that would still neglect Bob's so-far implicit reasoning
to the effect that $A^\prime$ is (potentially) not vulnerable to objection $B$.
If the dialogue continued from this point, detailed relationships between the constituent contents of `$A^\prime$' and `$B$' may need to be discussed, and an IATC analysis would be able to unpack these and account for the details.

\paragraph{In line with these design decisions, and inspired by the specific features of mathematical dialogue and exposition, IATC introduces a range of extra machinery to the IAT framework to model the relationships between mathematical objects and propositions, along with an array of dialogue moves related to the strategic aspects of proof.}
Unlike IAT, we make no attempt to cover argumentation in law, natural
science, or interpersonal mediation, fields in which the norms that
govern inference can be vastly different.  (Precedent, for example,
may be acceptable in a legal argument but not in one about ethics.)
In mathematical argumentation, many of the conventions are embodied in the objects under
discussion and the things that can sensibly be said about them.
Details of our notational apparatus are given in Tables
\ref{iatc-table:1} and \ref{iatc-table:2}.  
Appendix \ref{app:reference-coding-samples}
collects reference examples of short texts marked up with these codes.

\begin{table}[ht]
\begin{mdframed}
{\centering
\textbf{Performatives} \textbf{(}\texttt{perf[\ldots]}\textbf{)}

\par}

\smallskip

\begin{tabular}{@{\hspace{-.25ex}}p{.3\columnwidth}p{.65\columnwidth}}
\texttt{Assert} (\emph{s} [, \emph{a} ]) & Assert belief that statement \emph{s} is true, optionally because of \emph{a}.\\
\texttt{Agree} (\emph{s} [, \emph{a} ]) & Agree with a previous statement \emph{s}, optionally because of \emph{a}.\\
\texttt{Challenge} (\emph{s} [, \emph{a} ]) & Assert belief that statement \emph{s} is false, optionally because of \emph{a}.\\
\texttt{Retract} (\emph{s} [, \emph{a} ]) & Retract a previous statement \emph{s}, optionally because of \emph{a}.\\
\texttt{Define} (\emph{o}, \emph{p}) & Define object \emph{o} via property \emph{p}.\\
\texttt{Suggest} (\emph{s}) & Suggest a strategy \emph{s}.\\
\texttt{Judge} (\emph{s}, \emph{v}) & Apply a heuristic value judgement \emph{v} to some statement \emph{s}.\\
\texttt{Query} (\emph{s}) & Ask for the truth value of statement \emph{s}.\\
\texttt{QueryE} (\{$p_i$($X$)\}) & Ask for the class of objects \emph{X} for which all the properties $\{p_i\}$ hold.
\end{tabular}

\end{mdframed}
\caption{Inference Anchoring Theory + Content, part 1: Performatives\label{iatc-table:1}}
\end{table}

\newpage 
\paragraph{Our method for producing this set of tags was as follows.}
Two of us (with first degrees respectively in Mathematics and Information
Systems, both with more than 10 years experience studying
argumentation and social machines) performed close content analysis
\cite{klaus2004content} together on the first 100 comments in MiniPolymath
1.  Our analyses resulted in an initial tag set, including both typical
illocutionary performatives and mathematics specific performatives,
like \texttt{Define} and \texttt{QueryE}, as
needed (see Appendix \ref{app:reference-coding-samples} for examples).  Several
of the typical illocutionary connections (\texttt{Assert}, \texttt{Question}, \texttt{Challenge},
\texttt{Agree}) could be carried over from the schemes commonly applied in
IAT.  Our initial tag set was discussed and iteratively developed over
the same 100 comments by all co-authors, with any recurring
differences discussed, allowing us to align our results.  A third
co-author (with a first degree and PhD in Mathematics) then further
developed and refined the tag set by performing close content
analysis on the entire MiniPolymath 3 conversation and on sections of
MiniPolymath 1.  Again, this was conducted alongside discussion with
the other co-authors throughout the process.
A fourth co-author  (with a first degree in Mathematics)
later extended the tag set
with
additional informal logical relationships, such as \texttt{analogy}, and specific
content-focused relationships, such as \texttt{sums}, which played a
role in the further examples we treated in Section
\ref{iatc-examples}.  These extensions were again reviewed by all co-authors.

\begin{figure}[h]
\begin{center}
\includegraphics[trim=1.3cm 7cm 1mm 8.8cm,clip=true,width=0.9\textwidth]{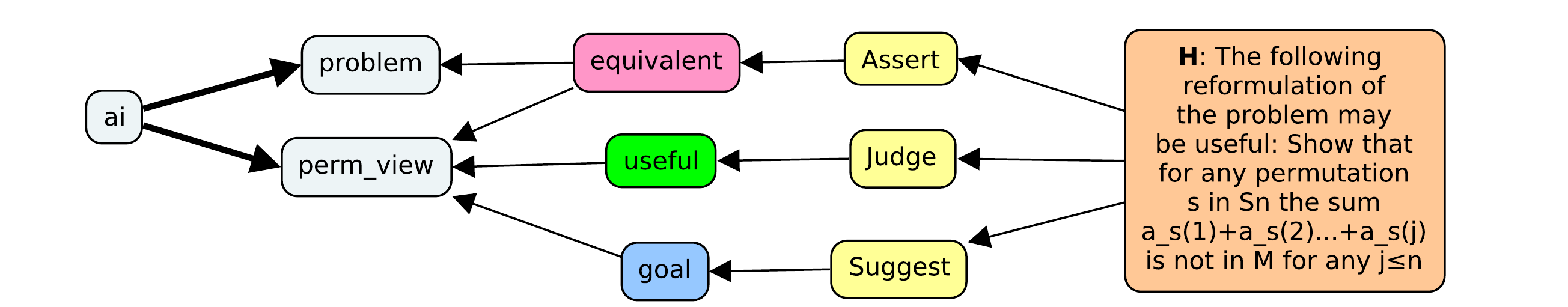}

\bigskip

\begin{minipage}{.8\textwidth}
\texttt{perf[\textcolor{perform}{Assert}](%
  rel[\textcolor{infer}{equivalent}](%
                   \textbf{\textcolor{gray}{problem}},
                   \textbf{\textcolor{gray}{perm\_view}}))}\newline
\texttt{perf[\textcolor{perform}{Judge}](%
       value[\textcolor{value}{useful}](%
       \textbf{\textcolor{gray}{perm\_view}}))}\newline
\texttt{perf[\textcolor{perform}{Suggest}](%
       meta[\textcolor{meta}{goal}](%
       \textbf{\textcolor{gray}{perm\_view}}))}\newline
\texttt{struct[used\_in](\textcolor{gray}{ai},
                        \textbf{\textcolor{gray}{problem}})}\newline
\texttt{struct[used\_in](\textcolor{gray}{ai},
                        \textbf{\textcolor{gray}{perm\_view}})}
\end{minipage}
\end{center}
\caption{IATC markup of the statement ``The following reformulation of the problem may be useful: Show that for any permutation $s$ in $S_n$, the sum $a_s(1)+a_s(2)\ldots +a_s(j)$ is not in $M$ for any $j\leq n$.''   A larger portion of the dialogue is analysed in graphical and textual form in Figure \ref{fig:running-example-analysed-graphic} and Table \ref{tab:running-example-analysed}. \label{fig:QuickIATCexample}}
\end{figure}

Our discussions concerned issues such as whether to label a statement such as
`\emph{it would be good to approach the problem in this way\ldots}'
as a simply a suggested \texttt{strategy} or, additionally, as a \texttt{value[...]} judgement
about the strategy.  Shortly, in Figure \ref{fig:QuickIATCexample}, we will
show an example tagging in which the multiple layers of interpretation are included.
However, perfect agreement about how to treat such
cases is not intended; the IATC framework is designed to account for
flexibility in interpretations.
The additional tags in Table \ref{iatc-table:2} were not at first
divided into the present categories, but repeated analysis quickly
revealed structural content relations, as well as inferential
structure, as natural categories, intuitively corresponding to the
mathematical and logical contents of the MiniPolymath discussions we examined.  By
far the most difficult categorisation to make was between value
judgements and reasoning tactics.  For example, the difference between
deeming a statement \texttt{useful} and suggesting it as a
\texttt{goal} could depend completely on how polite or how bold the
person making the utterance wished to be!
\begin{table}[ht]
\begin{mdframed}
{\centering
\textbf{Inferential Structure}  \textbf{(}\texttt{rel[\ldots]}\textbf{)}

\par}

\smallskip

\noindent
\begin{tabular}{@{\hspace{-.25ex}}p{.3\columnwidth}p{.65\columnwidth}}
\texttt{implies} (\emph{s}, \emph{t}) & Statement \emph{s} implies statement \emph{t}.\\
\texttt{equivalent} (\emph{s}, \emph{t}) & Statement \emph{s} implies statement \emph{t} and vice versa.\\
\texttt{not} (\emph{s}) & Negation of \emph{s}.\\
\texttt{conjunction} (\emph{s}, \emph{t}, \ldots) & Conjunction of statements \emph{s}, \emph{t}, \ldots \\
\texttt{has\_property} (\emph{o}, \emph{p}) & Object \emph{o} has property \emph{p}.\\
\texttt{instance\_of} (\emph{o}, \emph{m}) & Object \emph{o} is an instance of the broader class \emph{m}.\\
\texttt{indep\_of} (\emph{o}, \emph{d}) & Object \emph{o} does not depend on the choice of object \emph{d}.\\
\texttt{case\_split} (\emph{s}, \{$s_i$\}) & Statement \emph{s} is equivalent to the conjunction of the $s_i$'s.\\
\texttt{wlog} (\emph{s}, \emph{t}) & Statement \emph{t} is equivalent to statement \emph{s} but easier to prove.\\
\end{tabular}

\medskip

{\centering
\textbf{Heuristic Value Judgments} \textbf{(}\texttt{value[\ldots]}\textbf{)}

\par}

\smallskip

\noindent
\begin{tabular}{@{\hspace{-.25ex}}p{.3\columnwidth}p{.65\columnwidth}}
\texttt{easy} (\emph{s} [, \emph{t}]) & Statement \emph{s} is easy to prove; optionally, easier than statement \emph{t}.\\
\texttt{plausible} (\emph{s}) & Statement \emph{s} is plausible.\\
\texttt{beautiful} (\emph{s}) & Statement \emph{s} is beautiful (or mathematically elegant).\\
\texttt{useful} (\emph{s}) & Statement \emph{s} can be used in an eventual proof.\\

\end{tabular}

\medskip

{\centering
\textbf{Reasoning Tactics} \textbf{(}\texttt{meta[\ldots]}\textbf{)}

\par}

\smallskip

\noindent
\begin{tabular}{@{\hspace{-.25ex}}p{.3\columnwidth}p{.65\columnwidth}}
\texttt{goal} (\emph{s}) & Used with \texttt{Suggest} to guide other agents to work to prove the statement \emph{s}.\\
\texttt{strategy} (\emph{m}, \emph{s}) & Indicate that method \emph{m} might be used to prove \emph{s}.\\
\texttt{auxiliary} (\emph{s}, \emph{a}) & Statement \emph{s} requires an auxiliary lemma \emph{a}.\\
\texttt{analogy} (\emph{s}, \emph{t}) & Statement \emph{s} and statement \emph{t} should be seen as analogous in some way.\\
\texttt{implements} (\emph{s}, \emph{m}) & Statement \emph{s} implements the method \emph{m} from a previously suggested strategy.\\
\texttt{generalise} (\emph{m}, \emph{n}) & Method \emph{m} generalises method \emph{n}.  \\
\end{tabular}

\medskip

{\centering
\textbf{Content-Focused Structural Relations} \textbf{(}\texttt{struct[\ldots]}\textbf{)}

\par}

\smallskip

\noindent
\begin{tabular}{@{\hspace{-.25ex}}p{.3\columnwidth}p{.65\columnwidth}}
\texttt{used\_in} (\emph{o}, \emph{s}) & Object \emph{o} is used in statement (or object) \emph{s}.\\

\texttt{reform} (\emph{s}, \emph{t}) & Statement \emph{s} can be reformed into statement \emph{t}.\\

\texttt{instantiates} (\emph{s}, \emph{t}) & Statement \emph{s} schematically instantiates statement \emph{t}. \\
\texttt{expands} (\emph{x}, \emph{y}) & Expression \emph{x} expands to expression \emph{y}.\\
\texttt{sums} (\emph{x}, \emph{y}) & Expression \emph{x} sums to expression \emph{y}. \\
\texttt{cont\_summand} (\emph{x}, \emph{y}) & Expression \emph{x} contains \emph{y} as a summand. \\
\end{tabular}
\end{mdframed}
\caption{Inference Anchoring Theory + Content, part 2: Inferential Structure, Heuristics and Value Judgments, Reasoning Tactics, and content-focused relations\label{iatc-table:2}}
\end{table}
\FloatBarrier
Our performatives have \emph{slots}, which are filled by
\emph{statements} or \emph{objects}.  Statements may be represented in various ways:
in unparsed natural language, as symbolic tokens that serve as
shorthand for such statements, or in some representation language.
\paragraph{The other relations are clustered into segments treating \emph{Inferential Structure}, \emph{Heuristics and Value Judgments}, \emph{Reasoning Tactics}, and \emph{Content-Focused Structural Relations}.}
\paragraph{The associated grammatical categories are given the following abbreviations in our linear notation: `\texttt{rel}', `\texttt{value}', `\texttt{meta}', and \texttt{struct}'.}  For example, the expression
`\texttt{perf[Assert]}(\texttt{rel[has\_property]}(\emph{o},
\emph{p}))' denotes the assertion of the statement
``object \emph{o} has property \emph{p}.'' 
IATC allows direct, explicit, statements about objects, propositions, and statements.  For example, `\texttt{perf[Assert]}(\texttt{used\_in} (\emph{o}, \emph{s}))' denotes the assertion of the statement ``object $o$ appears in statement $s$.''

We have two notational strategies that call attention to features of discourse or content that are
taken as understood, but not explicitly stated.
Performatives may be marked as ``unspoken'' when the contents are only broadly implied.
Several examples of this notational strategy appear
in Section \ref{iatc-examples:gowers}.
Similarly, content-focused structural relations are sometimes introduced without an attached performative, whenever they have been noticed by the analyst.
Figure \ref{fig:QuickIATCexample} includes examples
of this latter usage.  This figure represents the analysis of a short
excerpt from  a real mathematical dialogue, showing its diagrammatic and textual representations in IATC.
The discussion (``MiniPolymath 1'') concerned Problem 6 from the 2009 International Mathematical Olympiad.  The text analysed in Figure \ref{fig:QuickIATCexample} is a portion of the fourth comment made in the discussion \cite[\href{https://wp.me/p3qzP-Ef\#comment-40141}{20 July, 6:50 am}]{tao2009imo}.
An expanded excerpt is discussed in Section \ref{iatc-examples:minipolymath}
along with more details of our IATC analysis of MiniPolymath data.
Here, colour coding highlights the correspondence between the
graphical and textual grammar elements.
One statement has been analysed into three performatives: 
\begin{itemize}\label{list:example-analysis}
\item The speaker \texttt{Assert}s that the problem has an \emph{equivalent} reformulation. ``The following \textbf{reformulation of the problem} may be useful\textbf{:} Show that \textbf{for any permutation $s$ in $S_n$, the sum $a_s(1)+a_s(2)\ldots +a_s(j)$ is not in $M$ for any $j\leq n$}.''
\item The speaker \texttt{Judge}s the reformulation to be (potentially) \emph{useful}. ``\textbf{The} following \textbf{reformulation} of the problem \textbf{may be useful}: [\ldots]''
\item The speaker \texttt{Suggest}s that the reformulation describes a \emph{goal} that could be worth pursuing:
``[\ldots] \textbf{Show that} [\ldots]''
\end{itemize}
In addition, mathematical objects (several symbols, $a_i$) are analysed
as component pieces of tagged content
(`\texttt{problem}' and `\texttt{perm\_view}').  Note that bold lines
at left in the figure are a shorthand for the `\texttt{used\_in}'
relation.  Subsequent statements in the dialogue will be able to link back to
these objects: the analysis of an expanded extract appears in Figure \ref{fig:running-example-analysed-graphic}.

\paragraph{The relations given in Tables \ref{iatc-table:1} and \ref{iatc-table:2} have been sufficient to describe the reasoning in a range of examples, however we do not claim that this list of relationships would treat all mathematical texts.}
Nor do these relationships describe mathematical texts at the
level of formality found in proof checking systems, or the level of
detail found in some other theorisations of discourse.
Thus, in the future IATC should not be limited to the set of tags presented here.
For example, we have found uses for the value judgments
`\texttt{easy}', `\texttt{beautiful}', and `\texttt{useful}', but it is
quite plausible that future work would find use for values such as
`\texttt{efficient}', `\texttt{generative}', or something else.
Similarly, useful additions may be found in the other grammatical categories.
The evidence from our examples in Section \ref{iatc-examples} is that
these major grammatical categories---performatives, inferential
relations, meta-level reasoning, value judgments, and content
relations---are themselves stable.

\paragraph{We have described, and illustrated with simple examples, the way content and strategic relationships can be used to mediate contextual relationships, but context is also representable in IATC in another more explicit way.}
Although IATC does not require proofs to be structured in a tree-like
hierarchy, nested structure is introduced as follows. 
In general, language elements in Table \ref{iatc-table:2}
that have a \emph{statement} slot can also have that slot filled by a
(possibly disconnected) \emph{subgraph}.  In this way, structure
corresponding to a ``lemma'' can be indicated.
A lemma, in this sense, is understood to be the reasoning that `\texttt{implements}' a
`\texttt{strategy}', or, alternatively, a specific section of reasoning
that `\texttt{implies}' some conclusion.   This representation strategy is similar to the ``partitioned
networks'' introduced by \citet{hendrix1975partitioned,hendrix1979encoding}.
An example will appear in Section \ref{iatc-examples:gowers}.

\paragraph{To summarise, IATC resembles IAT in many ways, but with changes
that are required when content, and discussions about content, are explicitly modelled.}
These features are necessary to express details of mathematical reasoning.
For example, one proposition that can be extracted from the statement in
Figure \ref{fig:QuickIATCexample} has the schematic form ``The reformulation $P$ is equivalent to the original question $Q$.''  IAT would have no way to extract $P$ and $Q$ 
from the assertion, but IATC can do so: they are
represented as `\texttt{problem}' and `\texttt{perm\_view}' in the figure.
Later moves can then connect to these pieces of content, and we
already see such structure forming in our analysis of the above short excerpt.

IATC retains and extends IAT's approach to modelling contents and
inferences, by adding non-propositional contents and more complex
logical and heuristic relations.  Illocutionary connections are also
retained, with some mathematics-specific additions.  However, IATC
sets aside the notion of transitions, not because we view dialogue
norms as unimportant, but because they are difficult to model at this
stage.  In IAT, relations between propositional contents roughly
mirror the norms involved.  The corresponding notion for IATC would be
heuristics that account for the production of new expressions, and
which take preceding expressions and background knowledge into
account.  We will have more to say about such heuristics in Section
\ref{iatc-examples}, nevertheless, many considerations must be
deferred to future work.

\section{Examples}\label{iatc-examples}

\thesis{In this section, we use three examples to showcase what IATC has to offer as a tool for analysis.}
We illustrate
\begin{itemize}
\item how IATC expresses the reasoning structures that arise in proof construction,
\item how it might be used to support computational models of mathematical reasoning,
\item and how it helps to uncover the salient elements of mathematical discourse.
\end{itemize}
To illustrate the points above, we have selected and analysed three examples
that exhibit informal, expository, and discursive features of mathematical reasoning.
The presentation here is a novel and self-contained synthesis and expansion of
remarks made in previous papers
\cite{corneli2017towards,corneli2017modelling,pease-and-martin}.
The three examples collectively show the richness of mathematical argument, and were selected
to match the three aims indicated above:
\begin{itemize}
\item Section \ref{iatc-examples:gowers}: A carefully spelled out informal solution to a tricky but non-technical mathematical problem serves to  illustrate the thought processes involved in successful mathematical problem solving.  The example shows how IATC captures this sort of thinking.
\item Section \ref{iatc-examples:mathoverflow}: A discussion of the relationships between, and merits of, different mathematical questions exhibits a level of abstraction above that needed in an individual proof.  We explore the ramifications for explicit representations of the reasoning involved.
\item Section \ref{iatc-examples:minipolymath}: A multi-participant dialogue that develops a challenging but not highly technical proof  casts light on processes of mathematical collaboration and mathematical reasoning.  An analysis of this material using IATC allows us to explore the process of proof-construction in detail.
\end{itemize}
In each of the following subsections, we give more details of the
context of each example, before presenting our analysis and comments.

\subsection{Making the reasoning explicit in the solution to a challenge problem}\label{iatc-examples:gowers}

In this section we aim to show that IATC is a natural modelling
tool for informal mathematics.
Whereas \citet[p.~23]{robinson1965machine} had sought to
\begin{quote}
reduce complex inferences, which are beyond the capacity of the human
mind to grasp as single steps, to chains of simpler inferences, each
of which is within the capacity of the human mind to grasp as a single
transaction,
\end{quote}
an alternative path of enquiry seeks to describe the heuristic process
of proving theorems in more cognitively plausible terms. 
In particular, one relevant question to ask is
how (human) mathematicians avoid large searches \cite{gowers-talk-ini}.
IATC can contribute to the further development of this effort,
by giving a uniform but expressive way to outline the
process of developing proofs.
Researchers working on mathematical software meant to exhibit
human-style reasoning may find this expressiveness useful.
\begin{figure}
\begin{mdframed}
What is the 500\textsuperscript{th} digit of $(\sqrt{2}+\sqrt{3})^{2012}$?

Even this, eventually, a computer will be able to solve.

For now, notice that total stuckness can make you do desperate things.
Furthermore, knowing the origin of the problem suggests good things to try.
The fact that it is set as a problem is a huge clue.

Can we do this for $(x+y)$? For $e$? Rationals with small denominator?

And how about small perturbations of these?  Maybe it is close to a rational?

$m$\textsuperscript{th} digit of $(\sqrt{2}+\sqrt{3})^n$?

$(\sqrt{2}+\sqrt{3})^2$?

$(2+2\sqrt{2}\sqrt{3}+3)$

$(\sqrt{2}+\sqrt{3})^2+(\sqrt{3}-\sqrt{2})^2=10$

$(\sqrt{2}+\sqrt{3})^{2012}+(\sqrt{3}-\sqrt{2})^{2012}$ is an integer!

And $(\sqrt{3}-\sqrt{2})^{2012}$ is a very small number.  Maybe the final answer is ``$9$''?

We need to check whether it's small enough. $(\sqrt{3}-\sqrt{2})^{2012}<\left(\frac{1}{2}\right)^{2012}=
\left(\left(\frac{1}{2}\right)^{4}\right)^{503}=
\left(\frac{1}{16}\right)^{503}<.1^{503}$, so we're in luck.

The answer is indeed $9$.
\end{mdframed}
\caption{A ``magic leap'' challenge problem and its solution, presented by Timothy Gowers as part of a public lecture at the University of Edinburgh, November 2, 2012.  (Reproduced from notes taken at the lecture.)\label{fig:magic-leap}}
\end{figure}

\paragraph{Our chosen example is a ``magic leap'' problem presented in a public lecture by Timothy Gowers, describing joint work with Mohan Ganesalingam \citeyearpar{gowers-talk}.}
The reasoning was communicated by a combination of speech and marks
on a chalkboard, and is reproduced in Figure \ref{fig:magic-leap}.
This example has been modelled in IATC by \citet{corneli2017modelling}.
The problem initially appears difficult to solve without computer algebra
system, but a simple algebraic solution is available
once the correct strategy is found.  As such, an
important part of the reasoning involved in solving the problem is to
find the correct strategy.  The steps involved in this part of the
reasoning process are heuristic rather than deductive.  We redescribe
the analysis here.

\begin{figure}
\includegraphics[width=\textwidth]{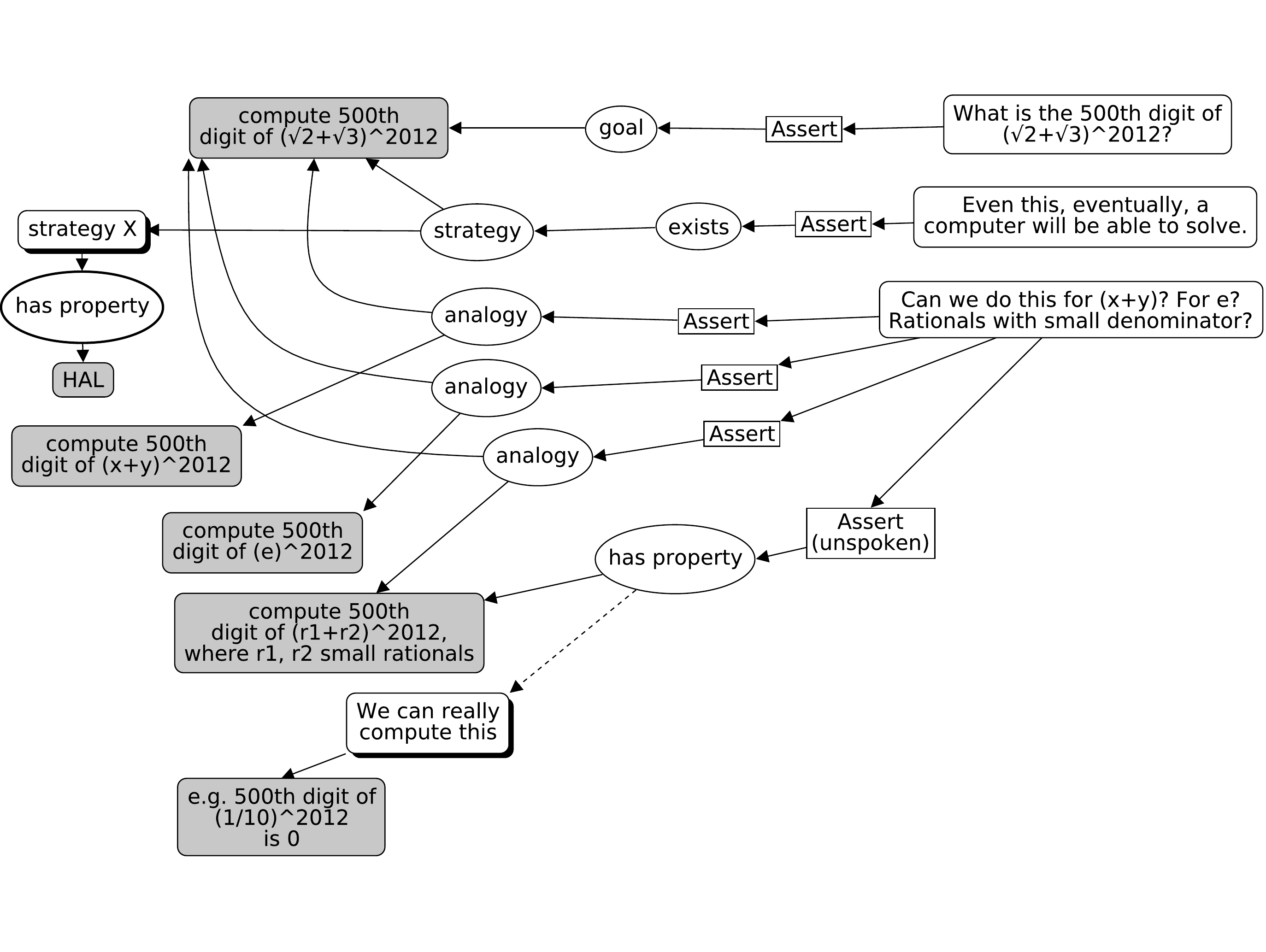}
\caption{IATC tagging for the first portion of the challenge problem\label{fig:challenge-problem-analogies}}
\end{figure}

\begin{figure}
\includegraphics[width=\textwidth,trim=9cm 15.5cm 9cm 0,clip=true]{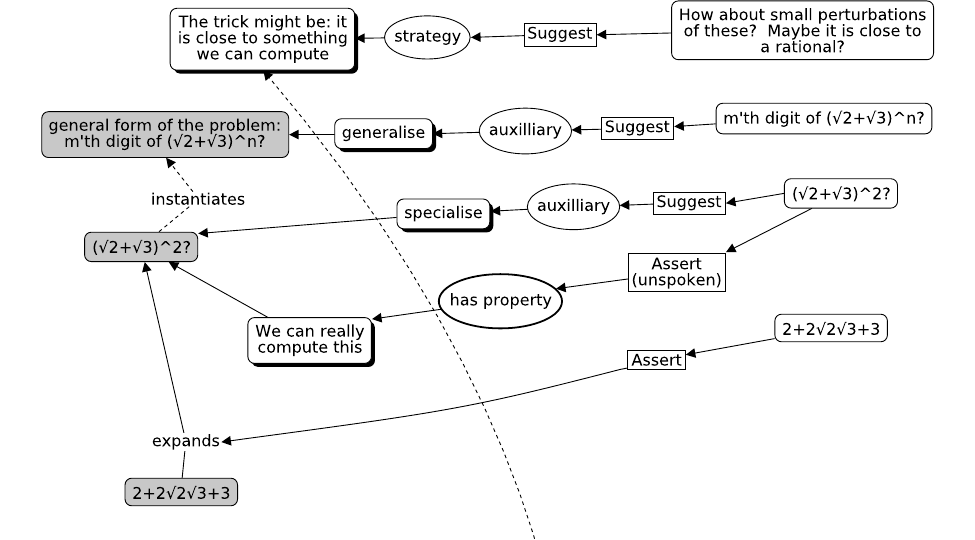}
\caption{IATC tagging for the second portion of the challenge problem\label{fig:challenge-problem-strategy}}
\end{figure}

\begin{figure}
\includegraphics[width=\textwidth,trim=0cm 8.5cm 0cm 0,clip=true]{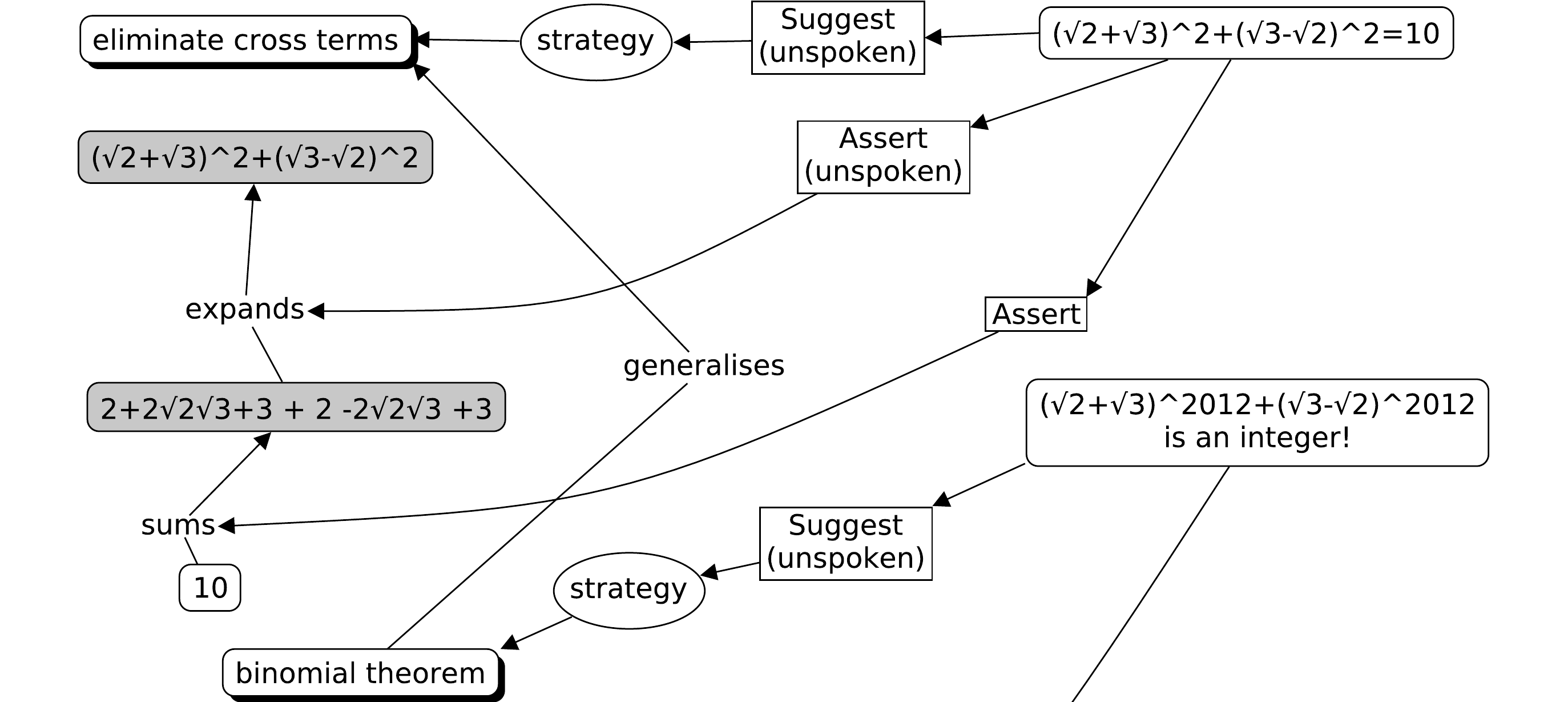}
\caption{IATC tagging for the third portion of the challenge problem\label{fig:challenge-problem-middle-bit}}
\end{figure}

\paragraph{For comparison with the IATC analysis, Figure \ref{fig:gowers2012-ala-lamport} reproduces the proof in Lamport's style. Figures \ref{fig:challenge-problem-analogies}, \ref{fig:challenge-problem-strategy}, \ref{fig:challenge-problem-middle-bit}
and \ref{fig:nested} present portions of the IATC tagging of the solution that was presented in Gowers's lecture.}
Figure \ref{fig:challenge-problem-analogies} illustrates an initial
exploration of the question, and Figure
\ref{fig:challenge-problem-strategy} establishes a `\texttt{strategy}'
based on that exploration (``The trick might be: it is close to
something we can compute'').
Figure \ref{fig:challenge-problem-middle-bit} opens the door to
applying the strategy.
The central part of the proof that `\texttt{implements}' the strategy
is highlighted in Figure \ref{fig:nested}.

\paragraph{The introduction to the proof, expanded in Figure \ref{fig:challenge-problem-analogies}---and condensed into a ``{\scshape Proof sketch}'' in Figure \ref{fig:gowers2012-ala-lamport}---contains interesting examples of heuristic reasoning.}
This part of the solution centres on the probing question ``Can we do
this for $\frak{X}$?'', where $\frak{X}$ ranges over several examples:
$x+y$, $e$, and small rationals, and where `this' denotes ``find the
500th digit of $\frak{X}^{2012}$.''  In the IATC representation, each
tentative proposal to ``do this\ldots'' stands in \texttt{analogy}
with the original problem statement.  Although Figure
\ref{fig:challenge-problem-analogies} contains only \texttt{Assert}
performatives, a more complete representation would also include
\texttt{Query} performatives, since the analogies are not only
proposed: their validity is also queried, much as we saw in the
example treated in the previous section.

Step \pfref{1} in the structured proof works out one of the ideas from the proof sketch at a
level of detail that was not present in the lecture, which instead
progressed directly on to the material treated in Step \pfref{2}.  As
\citet[p.~69]{fiedler2007argumentation} noted, ``The analysis of human
proof explanations shows that certain logical inferences are only
conveyed implicitly, drawing on the discourse context and default
expectations.'' There is no hard and fast rule that can tell us how
much of the implicit material we need to explicate, but one rule of thumb
that naturally arises from our representation strategy is that coherently
related discussions should correspond to connected graphs in the expansion.
Thus, for example, Figure \ref{fig:challenge-problem-analogies} includes an
implicit ``unspoken'' \texttt{Assert}ion; the proof is made fully
explicit in Step \pfref{1} of the Lamport-style proof, but never
appeared in the original lecture.  Again, in a standard
IAT representation, unspoken assertions would typically be represented as
`implicit' speech acts rooted on transitions, whereas in IATC, we see how
these unspoken assertions play a role in the argument via their expansion
and subsequent interconnections in the content layer.

\begin{figure}
\begin{mdframed}
\begin{minipage}{.95\textwidth}
\begin{proof}
\pfsketch\ What is the 500th digit of $(\sqrt{2}+\sqrt{3})^{2012}$?
Even
this, eventually, a computer will be able to solve.
The fact that this has been set as a problem is a huge clue.
Can we do this for
$x+y$? For $e$?  Small rationals?  And how about small perturbations
of these?  Maybe it is close to a rational?
\step{1}{\pflabel{1}For $n$ large enough and $m$ small enough in comparison, the
  \(m\)th digit of a sufficiently small rational $r$ to the \(n\)th power
  is equal to 0.}
        \begin{proof}
        \pf\
        \case {$r< 1/10$}
        \step{1.1}{$(1/10)^n=.1^n$ has 0 in $n-1$ places in its decimal expansion.}
        \step{1.2}{$r<1/10$ implies $r^n$ has zeros in at least $n-1$ places in its decimal expansion, so we simply need to select $m<n$. \qed}

        \end{proof}
\step{2}{\pflabel{2}Can we compute the \(m\)th digit of $(\sqrt{2}+\sqrt{3})^n$?}
         \begin{proof}
         \pf\
         \case {$n=2$}
         \step{2.1}{\pflabel{2.1}$(\sqrt{2}+\sqrt{3})^2=2+2\sqrt{2}\sqrt{3}+3$ {\Large $\ast$}}
         \end{proof}
\step{3}{\pflabel{3}Step \pfref{2.1} fails to give us an answer by direct
  computation, but if we eliminate cross-terms, we can see that
  $(\sqrt{2}+\sqrt{3})^2$ is ``close to'' an integer.}
  \begin{proof}
  \pf\ \step{3.1}{\pflabel{3.1}$(\sqrt{2}+\sqrt{3})^2 + (\sqrt{2}-\sqrt{3})^2 = 10$}
       \step{3.2}{$(\sqrt{2}+\sqrt{3})^2 = 10-(\sqrt{2}-\sqrt{3})^2$}
       \step{3.3}{\pflabel{3.3}$(\sqrt{2}-\sqrt{3})^2$ is small}
        \begin{proof}
        \pf\ \step{3.3.1}{\pflabel{3.3.1}$(\sqrt{2}-\sqrt{3})^2 \approx (1\frac{1}{3} - 1\frac{1}{2})^2$ by continued fraction approximation.}
             \step{3.3.2}{$(1\frac{1}{3} - 1\frac{1}{2})^2=(\frac{8}{6}-\frac{9}{6})^2 = 1/36$ \qed}
        \end{proof}
  \end{proof}
\step{4}{\pflabel{4}Generalising Step \pfref{3.1}, $(\sqrt{2}+\sqrt{3})^{2012} +
  (\sqrt{2}-\sqrt{3})^{2012}$ is an \emph{integer}!}
\begin{proof}
  \pf\ \step{4.1}{Terms with odd exponents cancel by the binomial theorem. \qed}
  \end{proof}
\step{5}{\pflabel{5}And $(\sqrt{2}-\sqrt{3})^{2012}$ is a \emph{very} small number. Maybe the final answer will be ``9''? We need to check whether it's small enough.}
\begin{proof}
  \pf\ \step{5.1}{\pflabel{5.1}$(\sqrt{2}-\sqrt{3})^{2012} < (1/2)^{2012}$ by Step \pfref{3.3.1}}
       \step{5.2}{$(1/2)^{2012}=((1/2)^4)^{503}$}
       \step{5.3}{$((1/2)^4)^{503}=(1/16)^{503}$}
       \step{5.4}{\pflabel{5.4}$(\sqrt{2}-\sqrt{3})^{2012}<(1/16)^{503}<.1^{503}$, and so has at least 502 zeros in its decimal expansion by Step \pfref{1}}
       \step{5.5}{\pflabel{5.5}The answer is indeed 9. \qed}
  \end{proof}
\end{proof}
\end{minipage}
\end{mdframed}
\caption{The solution to the challenge problem as a Lamport-style
  structured proof.\label{fig:gowers2012-ala-lamport}}
\end{figure}

\paragraph{Indeed, nowhere in the explicitly communicated reasoning is the key \texttt{strategy} fully and explicitly stated.}
The basic strategy of the proof is that \emph{the quantity of interest
  may be sufficiently close to something we can compute}.  In the IATC
representation (Figure \ref{fig:challenge-problem-strategy}), this is
understood to be \texttt{Suggest}ed by the following statements from
the proof sketch, ``And how about small perturbations of these?  Maybe
it is close to a rational?''  Step \pfref{1} of the structured proof
shows that rationals do, in fact, match the strategy's preconditions.
The IATC representation is less explicit on this point, since it
sticks more closely to the reasoning expressed in the lecture.  This
example shows that even relatively explicit statements may need
further interpretation to be represented meaningfully in IATC.
Specifically, the way the proof progresses only makes sense if we
recognise the `\texttt{strategy}' implied by what might otherwise
appear to be a throwaway comment early on.

\paragraph{Step \pfref{2} in the structured proof concerns another analogy.}
This time, a special one which, the IATC analysis notes,
symbolically \texttt{generalise}s the initial question (Figure
\ref{fig:challenge-problem-strategy}).  
That is, rather than considering $(\sqrt{2}+\sqrt{3})^{2012}$ we
now consider $(\sqrt{2}+\sqrt{3})^{m}$.
(NB.~an edge connecting the  `\texttt{generalise}' node to the problem statement has been omitted.)
However, the concept of generalisation remains implicit in the
corresponding portion of the structured proof.  Indeed, Step \pfref{2}
is not a good match for the requirements of structured proof at all,
since it is not a real lemma, and its ``proof'' fails (indicated by
``{\Large $\ast$}'').  Including failed proof steps is not a problem
for IATC.  In Figure \ref{fig:challenge-problem-middle-bit} the
process of solving the problem proceeds apace, without pausing to
remark on a failed lemma, now that something more interesting has been
discovered.

\paragraph{Meanwhile, Step \pfref{3} in the structured proof implements the main strategy for resolving a special case of our generalised problem, namely showing that $(\sqrt{2}+\sqrt{3})^{2}$ is close to an integer, establishing a pattern that leads to the conclusion.}
Again, Step \pfref{3.3} offers considerably more detail than was present in the
original lecture.

Step \pfref{4} subsequently generalises the
method that was used in Step \pfref{3}, and applies it to the
expression we were originally interested in.
Figure \ref{fig:nested} diagrams out the reasoning that underlies this step.
The long-range dashed edge in this figure connects with the
node ``The trick might be: it is close to something we can compute''
pictured in Figure \ref{fig:challenge-problem-strategy}.
The
collection of nodes highlighted in red \texttt{implement} that
strategy.  Notice, though, that the computation is not done explicitly:
it's unimportant which integer the number of interest is close to.
Collectively, the fact that
$(\sqrt{2}+\sqrt{3})^{2012}+(\sqrt{3}-\sqrt{2})^{2012}$
sums to ``some integer'' and the fact that $(\sqrt{3}-\sqrt{2})^{2012}$
is sufficiently small \texttt{implies} the result.
Step \pfref{5} shows the details of the final computational check.

\paragraph{Several objections could be raised about the structured proof presented in Figure \ref{fig:gowers2012-ala-lamport}, most notably to the inclusion of a failed lemma in Step \pfref{2}.}
However, as a source of information about the intuition behind the
proof, this failure is valuable.  While objections to the IATC treatment 
are also possible, it is clear that this method helps to 
make explicit features of the proof process that remain implicit in the
structured proof.  In particular,
analogies, strategies, and relationships between methods are
made explicit.  While the structured proof augments the lecture with
more technical details, IATC provides a more faithful model of the reasoning
expressed in the lecture itself.

{
\renewcommand{\bottomfraction}{.9}
\begin{figure}[t]
{\centering
\begin{tikzpicture}
    \node[anchor=south west,inner sep=0] (image) at (0,0) {\includegraphics[width=.8\textwidth,trim=9cm 6.5cm 8.5cm 7.5cm,clip=true]{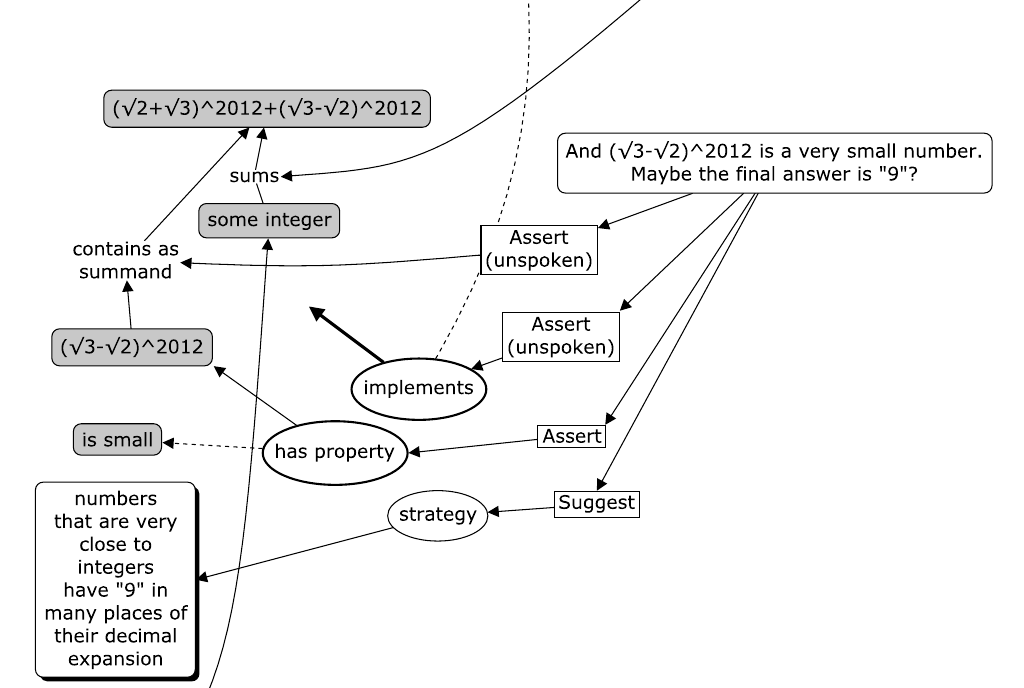}};
    \begin{scope}[x={(image.south east)},y={(image.north west)}]
\draw [red,ultra thick] plot [smooth cycle] coordinates {(0.07,0.32) (0.03,.5) (0.1,.87) (.25,.92) (.4,.9) (.44,.86) (.42,.8)  (.3,.56) (.2,.35)};
        \draw[blue,ultra thick,rounded corners] (0.187,0.645) rectangle (0.338,0.713);

    \end{scope}
\end{tikzpicture}
\par}
\caption{Nested structure (in red) \texttt{implements} the \texttt{strategy} suggested earlier:
``The trick might be: it is close to
something we can compute.''  The intermediate conclusion reached in this phase of reasoning (highlighted in blue), when taken together with a further computational check, subsequently \texttt{implies} that the answer is ``9''. \label{fig:nested}}
\end{figure}
}

\FloatBarrier
\subsection{Towards computable models of mathematical reasoning via IATC: A Q\&A example}\label{iatc-examples:mathoverflow}

Contributors to discussions about mathematics on MathOverflow do more than just talk about proofs.
\begin{quote}
The presentation is often speculative and informal, a style which
would have no place in a research paper, reinforced by conversational
devices that are accepting of error and invite
challenge. \cite{martin2013does}
\end{quote}
\paragraph{IATC allows the argumentation aspects of mathematical dialogues to be represented as explicit graphical structures, which gives a plausible basis from which to develop an explicit computational model of the reasoning steps that are implied in mathematical argumentation.}
\paragraph{\citet{corneli2017towards} showed how IATC could be used to create graphical models of the discussion that develops around a question posted on MathOverflow.}  Here we will remark further on implications for computational modelling.
The question, which was given the title  ``Group cannot be the union of conjugates'' \cite{chandrasekhar2010group} is as follows:

\medskip

\noindent ``\emph{I have seen this problem, that if $G$ is a finite group and $H$ is a
  proper subgroup of $G$ with finite index then $ G \neq\hspace{-.2em}\bigcup\limits_{g \in G} gHg^{-1}$. Does this remain true for the
  infinite case also?}''

\noindent 
In the most straightforward reading, two superficially similar group-theoretic propositions seem
to be at stake:
\begin{enumerate}[label=(\emph{P\arabic*}),ref=(\emph{P\arabic*}),leftmargin=1cm]
\item ``\emph{If $G$ is a \uline{finite} group, $H$ is a subgroup of $G$ and the
  index $[G \mathop{:} H]$ is finite, then $G$ is not equal to the
  union of $gHg^{-1}$}''; and, \label{qa:lhs:prop}
\item ``\emph{If $G$ is an \uline{infinite} group, $H$ is a subgroup of $G$
  and the index $[G \mathop{:} H]$ is finite, then $G$ is not equal to
  the union of $gHg^{-1}$}.'' \label{qa:rhs:prop}
\end{enumerate}
The question thus implicitly outlines an argument by analogy:
\begin{itemize}
\item \ref{qa:lhs:prop} is true
\item \ref{qa:rhs:prop} is similar to \ref{qa:lhs:prop}
\item Therefore, \ref{qa:rhs:prop} is (potentially) true as well
\end{itemize} 
The essence of the question is to ask whether the mathematical facts
align with this schematic argument.
As it turns out, this question is answered in the affirmative.
Shortly after the question was asked, one discussant make the terse
comment ``the case of infinite $G$ readily reduces to the case of
finite $G$''; months later, another discussant supplies an explicit
proof of \ref{qa:rhs:prop}.

In the mean time, other discussants had proposed and addressed several
alternative formulations of the question.  An important distinction hinges on the
interpretation of the phrase ``\emph{infinite case}.''
An alternative proposition that incorporates some of the suggested
revisions is as follows:
\begin{enumerate}[start=2,label=(\emph{P\arabic*}$^\prime$),ref=(\emph{P\arabic*}$^\prime$),leftmargin=1cm]
\item ``\emph{If $G$ is an \uline{infinite} group, $H$ is a proper finite index subset of $G$
  and the index $[G \mathop{:} H]$ is \uline{infinite}, then $G$ is not equal to
  the union of $gHg^{-1}$}.'' \label{qa:rhs:revised}
\end{enumerate}
In this case an argument by analogy would not match the facts:
a counterexample is supplied to show that proposition \ref{qa:rhs:revised} is false.

The dialogue is an interesting example of mathematical reasoning in
which proof certainly plays a role, but is nevertheless of secondary
interest compared with asking interesting questions, and thinking
about how different questions relate to each other.
What would be necessary to represent this sort of dialogue computationally?
Expressing propositions like \ref{qa:lhs:prop} in IATC is 
straightforward, though, as we noted, the content layer is not
directly modelled in this representation language.  The
following expression represents this proposition in IATC, introducing
additional invented pseudocode representations (in italics) in the content layer.
\begin{Verbatim}[commandchars=\\\{\}]
perf[Assert](
 rel[implies](
  rel[conjunction](\emph{finite_group}(G),
                   \emph{subgroup}(H,G),
                   rel[has_property](\emph{index}(H,G),\emph{is_finite}))
  rel[not](\emph{equal}(G,
                 \emph{union_over}(\emph{conjugates}(H,g),\emph{elements}(g,G))))))
\end{Verbatim}

Processing such expressions to build a model of a dialogue
will require adding numerous stanzas like this one, each rooted on an IATC
performative, into one graph database that records the relationships
between the statements and their constituent parts.  Individual expressions
like the \texttt{implies} relationship would need to be addressable,
in order for an \texttt{analogy} between two implications to be
proposed.
Definitions for predicates like \texttt{\textit{finite_group}}
and special constructions like \texttt{\textit{union_over}}
could be supplied in an accompanying knowledge base.   In further rounds of
computational processing, the analogies between
\ref{qa:lhs:prop} and \ref{qa:rhs:prop}, and between
\ref{qa:lhs:prop} and \ref{qa:rhs:revised}, could be checked
 using graph-processing methods described
by \citet{sowa2003analogical}.  
New heuristics would be needed
if the aim was to demonstrate the truth or falsity of the various
propositions, not just to recreate the surface analogies.
Moreover, as we've seen, mathematical dialogues are not just concerned
with verifying statements, but may
also consider the qualities that make a particular
question interesting in a given context.   Heuristics that can 
be used to select interesting problems are not prevalent in current 
mathematical software.

As a limited proof of concept showing the plausibility of adding a
computational deduction and verification layer on top of IATC
representations, \citet{corneli2017modelling} give a detailed
expansion of one step of a mathematical proof using simple rules
for transforming the underlying graph structures.
It is worth emphasising that the representations
of reasoning afforded by language elements in Tables
\ref{iatc-table:1} and \ref{iatc-table:2} do not themselves
encode the meta-level reasoning associated with such graph transformations.
\subsection{MiniPolymath Revisited}\label{iatc-examples:minipolymath}

The data that underlie this section were generated in a series of online experiments in collaborative problem solving convened by mathematician Terence Tao \citeyearpar{tao2009imo,tao2011imo}.
\thesis{We use IATC to expand on a previous analysis of
  this data presented by \citet{pease-and-martin},} showing how IATC can
advance the theory of mathematical argument through the
detailed analysis of real world examples, as per \citet{Carrascal2015}.

In their 2012 paper,
Pease and Martin analysed the third MiniPolymath project in broad strokes,
with each blog comment comprising a single unit to be tagged.
They developed a typology of five intuitive comment types, based on
the mathematical content of each comment: \emph{examples},
\emph{conjectures}, \emph{concepts}, \emph{proofs}, and \emph{other}.

In order to assign comments to these categories, both authors
performed close content analysis, together, on all comments posted
between the time at which Tao posted the problem to his blog
(8pm, UTC on July 19th, 2011) and the time he announced
that a solution had appeared (9.50pm, UTC on July 19th, 2011).
The discussion comprised 147 comments over 27 threads.

Ten comments were assigned to more than one category.
(Both authors have a first degree in
Mathematics; one of us has a PhD in Mathematics and over 10 years
experience as a professional research mathematician; the other has a
PhD in a related discipline and more than 10 years experience studying
mathematical reasoning.)

Our present IATC analysis of the same data is designed to give a more complete picture of the
linguistic, dialectical, and inferential structure of the comments
that fall within the five intuitive categories mentioned.  There are three main
differences between the two analyses.  First, in comparison withh the earlier
broad-stroke analysis, the IATC analysis is richly detailed, with a
unit defined as any quantum of commentary with taggable
content.  Secondly, our focus in the earlier analysis was purely on
mathematical content, and on the {\em type} of mathematical
content in particular.  This contrasts with our
present analysis, in which we provide a more fine-grained representation of mathematical
content in the taggable units, and furthermore take into account
linguistic, dialectical, and inferential structure.
Third, the IATC analysis takes into consideration the entire
MiniPolymath 3 conversation, including the comments that came after
Tao had announced that a proof had been found.

The new analysis, accordingly, adds depth to our earlier analysis.
Crucially, the new perspective will be more relevant to argumentation
theorists, and supports a detailed understanding of what went on in
the process of constructing the collaborative proof.  The earlier
typology provided an initial way to sort the content, whereas the IATC
tag set developed along with our analysis via the iterative,
discursive method discribed in Section \ref{iatc}.  Though they cover
the same data and show some correlations, as described below,
the latter categorisation was not derived from the earlier one.

\begin{figure}
\begin{center}
\includegraphics[width=.62\textwidth]{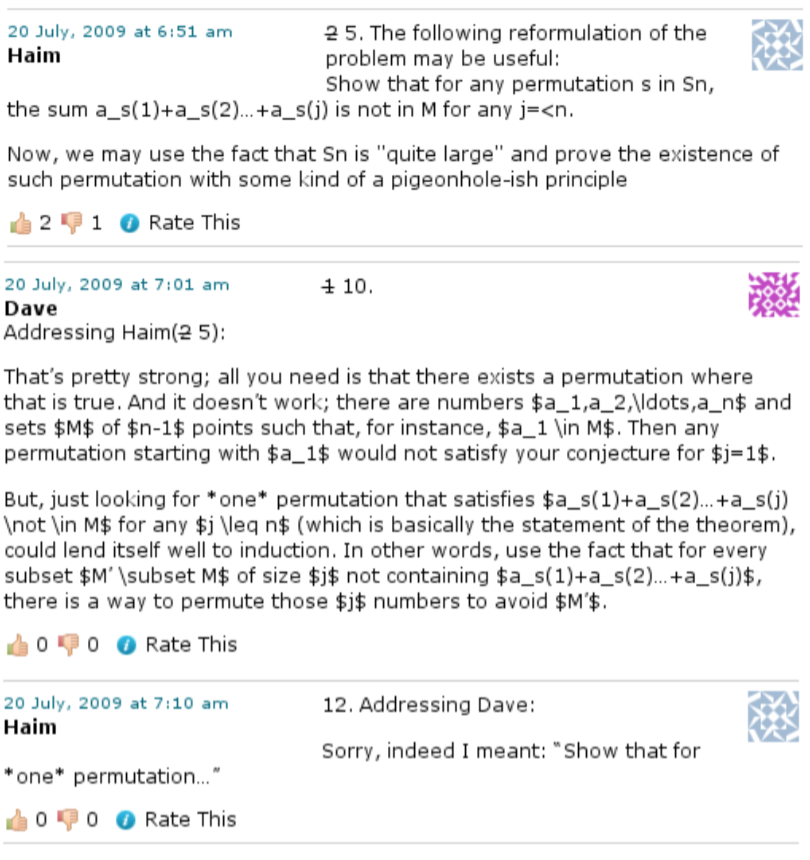}
\end{center}
\caption{Screenshot of a portion of the MiniPolymath 1 dialogue\label{fig:running-example-introduced}}
\end{figure}

\begin{figure}[h]\label{IATCtagging}
\begin{center}
\includegraphics[trim=2mm 0 1mm 18mm,clip=true,width=0.9\textwidth]{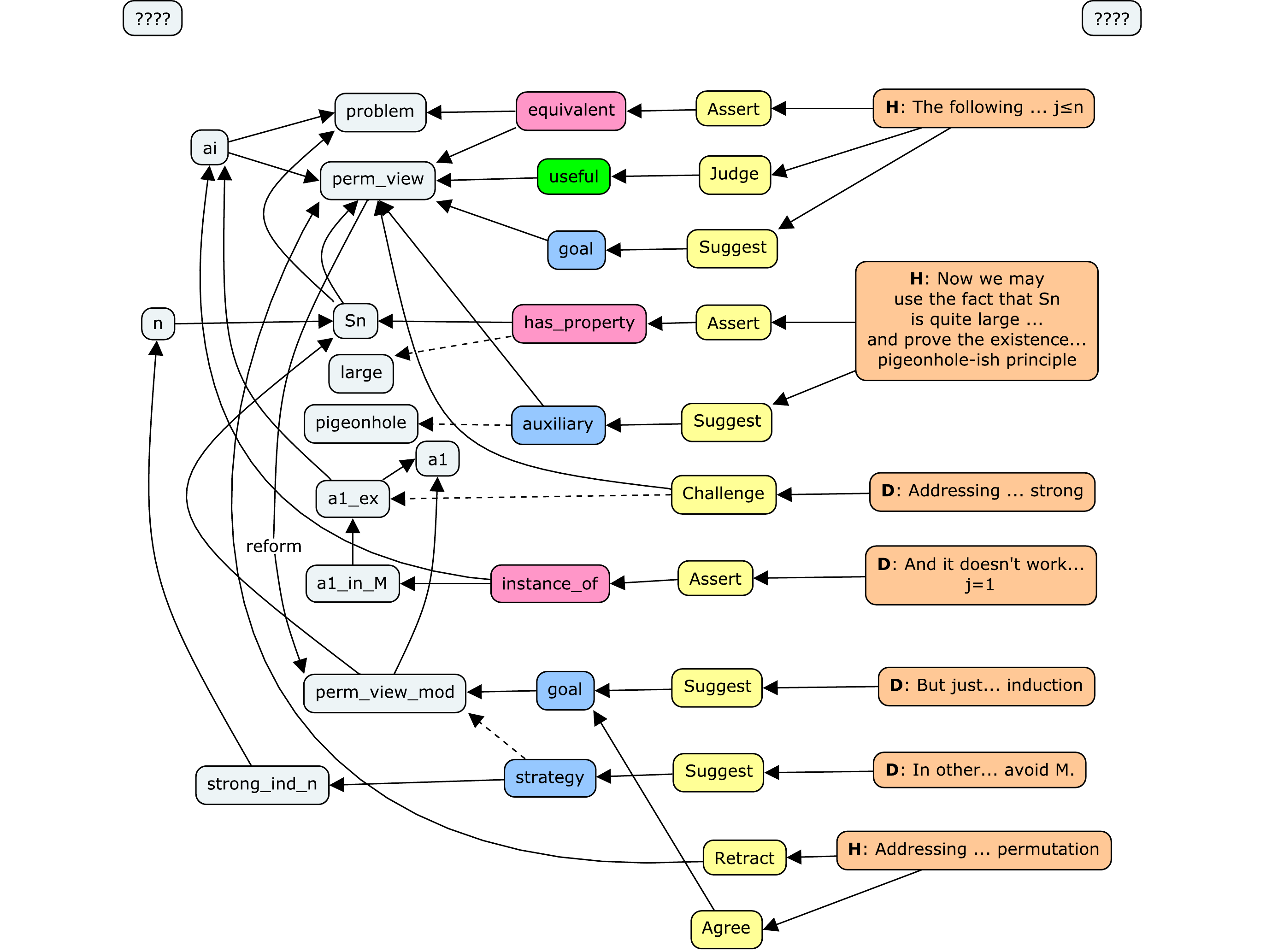}
\end{center}
\caption{IATC analysis of MPM1 excerpt (graphical form)\label{fig:running-example-analysed-graphic}}
\end{figure}

\paragraph{Figure \ref{fig:running-example-introduced} presents an excerpt from the MiniPolymath 1 dialogue (MPM1) as it originally appeared on Tao's blog.}
Figure \ref{fig:running-example-analysed-graphic} and Table
\ref{tab:running-example-analysed} give the IATC analysis of this
excerpt in diagrammatic and textual form.  The first portion of Figure
\ref{fig:running-example-analysed-graphic} repeats the contents of Figure
\ref{fig:QuickIATCexample}.  The longer excerpt shown here illustrates
complex contextual interconnections forming in the content
layer.

\begin{figure}[h]
\begin{center}
\includegraphics[trim=3cm 1.2cm 3cm 1.2cm,clip=true,width=\textwidth]{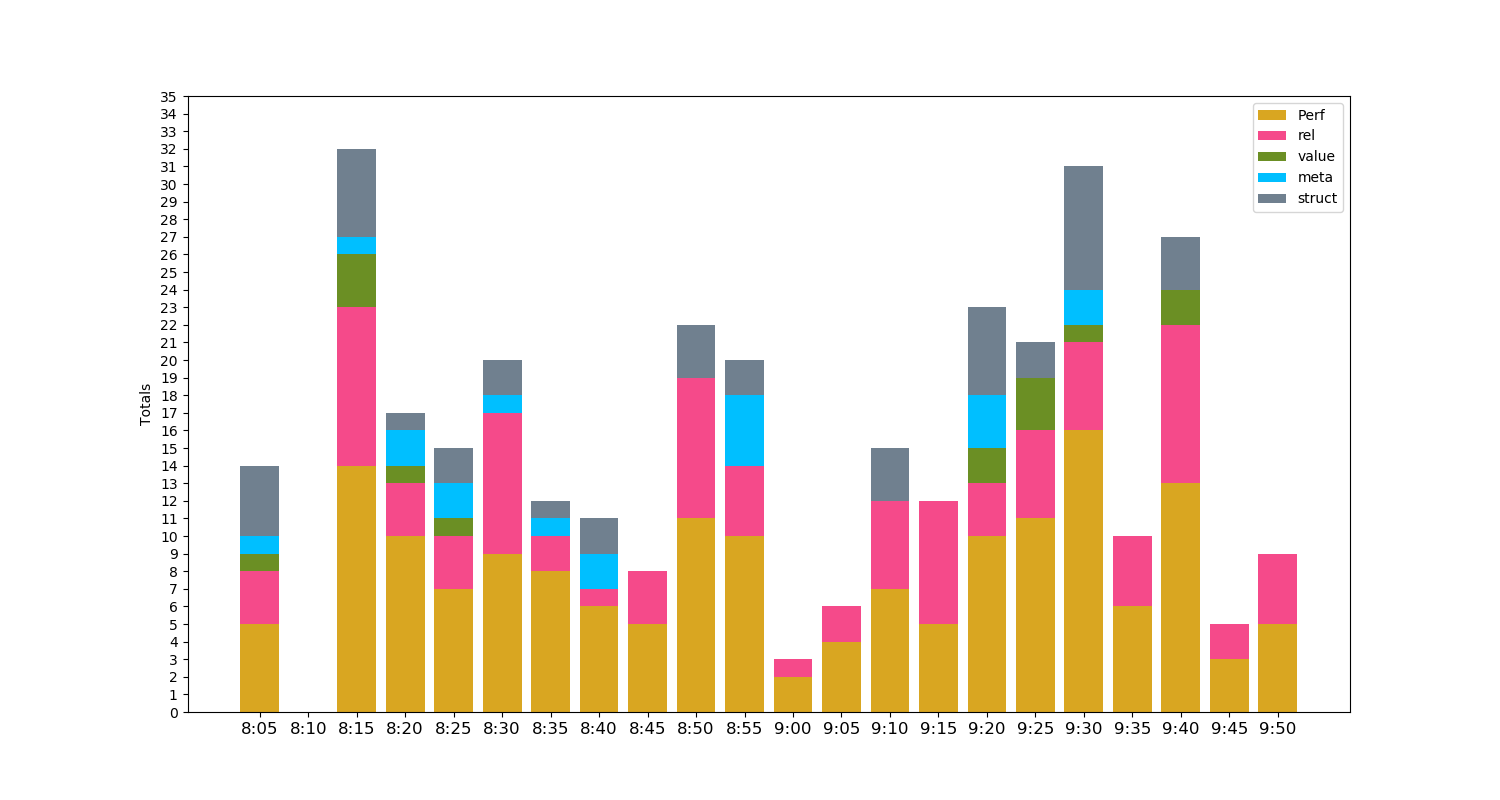}
\end{center}
\caption{Timeline of the MiniPolymath 3 dialogue showing the IATC grammar categories used in the tagging.   Comments are binned into 5 minute intervals.  The first interval is 8:05-8:09 and the last is 9:50-9:59, inclusive.
\label{fig:grammar-categories}}
\end{figure}

\begin{figure}[h]
\begin{center}
\includegraphics[trim=3cm 1.2cm 3cm 1.2cm,clip=true,width=\textwidth]{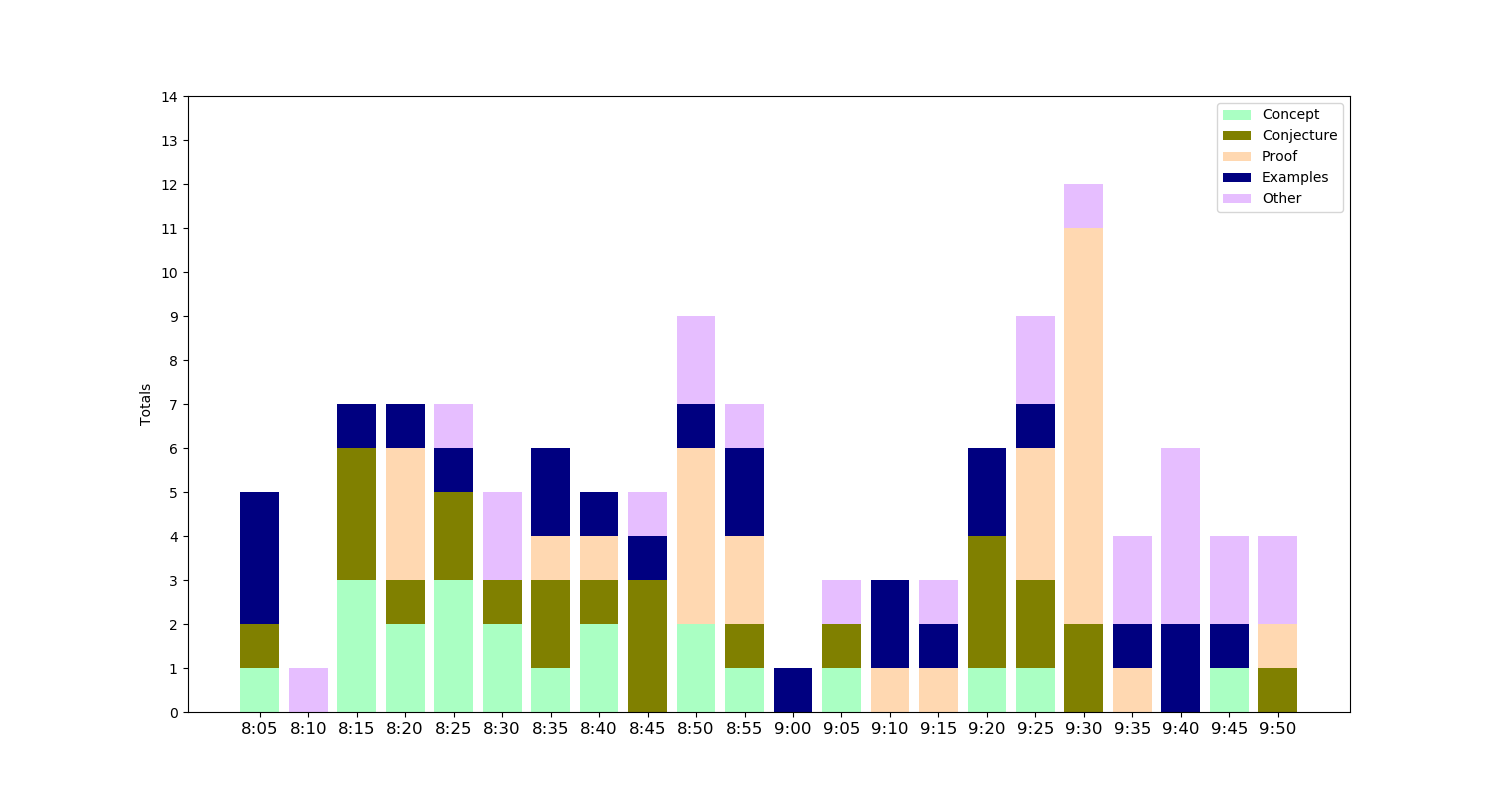}
\end{center}
\caption{Timeline of the MiniPolymath 3 dialogue showing comments categorised into five categories: Concept, Conjecture, Proof, Examples, and Other.
\label{fig:alternative}}
\end{figure}

\paragraph{Our main example in this section is MiniPolymath 3 (MPM3),
which we tagged into IATC in its entirety.}
(This work was carried out by one co-author with a first degree and PhD in Mathematics,
in consultation with others as described in Section \ref{iatc}.)
As an indicative sample, the first three comments and their tags are shown in Figure
\ref{fig:mpm3-example}.
Figure \ref{fig:grammar-categories} shows how tags from IATC's five grammatical
categories were distributed over time.  Thus, for example, we see
`\texttt{value}' tags used early in the discussion as strategies are
being considered, and again later in the discussion when solutions are
being vetted.
Figure \ref{fig:alternative} gives another view of the timeline,
showing how the comments were categorised into the 5-part typology
from Pease and Martin.  In the initial
categorisation developed for that paper, comments were allowed to be
in multiple categories at once.
\paragraph{Here, to facilitate a clean mapping to IATC, we redid the categorisation with the requirement that each comment should fit into exactly one main category.}
We arrived at a nearly equal division of comments among the five categories:
\emph{example} (20.3\
\emph{concept} (19.5\
(19.5\
one of the coauthors with a first degree in Mathematics.)

Figure \ref{fig:piecharts} illustrates the correspondence between IATC
tags with the earlier typology.  Aligning the bulkier 5-part
categorisation with the IATC tagging shows that these five intuitive
labels are mapped in very different ways to the more detailed IATC tag
set.

\begin{table}[t]
\setcounter{turn}{0}
{\centering\fontsize{9.5}{11}
\begin{tabular}{!{\vrule width 1pt} p{.95\textwidth} !{\vrule width 1pt}}
\bottomrule
(\textbf{Haim}) The following reformulation of the problem may be useful:
``Show that for any permutation $s$ in $S_n$, the sum
$a_{s(1)}+a_{s(2)}+\ldots +a_{s(j)}$ is not in $M$ for any $j\leq n$.''\\

\vspace{-.5\baselineskip}
\newturn
\texttt{perf[\textcolor{perform}{Assert}](%
  rel[\textcolor{infer}{equivalent}](%
                   \textbf{\textcolor{gray}{problem}},
                   \textbf{\textcolor{gray}{perm\_view}}))},\newline
\newturn
\texttt{perf[\textcolor{perform}{Judge}](%
       value[\textcolor{value}{useful}](%
       \textbf{\textcolor{gray}{perm\_view}}))},\newline
\newturn
\texttt{perf[\textcolor{perform}{Suggest}](%
       meta[\textcolor{meta}{goal}](%
       \textbf{\textcolor{gray}{perm\_view}}))},\newline
\newturn
\texttt{struct[used\_in](\textcolor{gray}{ai},
                        \textbf{\textcolor{gray}{problem}})},\newline
\newturn
\texttt{struct[used\_in](\textcolor{gray}{ai},
                        \textbf{\textcolor{gray}{perm\_view}})}\\  \hline
(\textbf{---}) Now, we may use the fact that $S_n$ is ``quite
large'' and prove the existence of such
permutation \\
\vspace{-.5\baselineskip}
\newturn
\texttt{perf[\textcolor{perform}{Assert}](%
      rel[\textcolor{infer}{has\_property}](\textcolor{gray}{Sn}, \textcolor{gray}{large}))},\newline
\newturn
\texttt{struct[used\_in](\textcolor{gray}{Sn},
                        \textbf{\textcolor{gray}{perm\_view}})}, \newline
\newturn
\texttt{struct[used\_in](\textcolor{gray}{Sn},
                        \textbf{\textcolor{gray}{problem}})}\\  \hline
(\textbf{---}) with some kind of a pigeonhole-ish
principle. \\
\vspace{-.5\baselineskip}
\newturn
\texttt{perf[\textcolor{perform}{Suggest}](%
        meta[\textcolor{meta}{auxiliary}](%
        meta[\textcolor{meta}{strategy}](\newline
       \phantom{indentat}\textbf{\textcolor{gray}{perm\_view}},%
       \textbf{\textcolor{gray}{pigeonhole}}),
       rel[\textcolor{infer}{has\_property}](\textcolor{gray}{Sn}, \textcolor{gray}{large})))}\\ \toprule \bottomrule

(\textbf{Dave}) Addressing Haim(2 5):
That's pretty strong. And it doesn't work; there are numbers
$a_1,a_2,\ldots,a_n$ and sets $M$ of $n-1$
points such that, for instance, $a_1 \in M$.
Then any permutation starting with $a_1$
would not satisfy your conjecture for $j=1$.\\
\vspace{-.5\baselineskip}
\newturn
\texttt{perf[\textcolor{perform}{Challenge}](%
     \textbf{\textcolor{gray}{perm\_view}}, 
     \textbf{\textcolor{gray}{a1\_ex}})},\newline
\newturn
\texttt{perf[\textcolor{perform}{Assert}](%
         rel[\textcolor{infer}{instance\_of}](%
            \textcolor{gray}{a1}, \textcolor{gray}{a1\_in\_M}))},\newline
\newturn
\texttt{struct[used\_in](\textbf{\textcolor{gray}{a1\_ex}}, \textcolor{gray}{a1\_in\_M})}\\ \hline

(\textbf{---}) But, just looking for {\bf one} permutation that
satisfies $a_{s(1)}+a_{s(2)}+ \ldots +a_{s(j)} \not \in M$ for
any $j \leq n$ (which is basically the statement
of the theorem), could lend itself well to induction.\\
\vspace{-.5\baselineskip}
\newturn
\texttt{perf[\textcolor{perform}{Suggest}](meta[\textcolor{meta}{goal}](perm\_view\_mod))},\newline
\newturn
\texttt{struct[reform](\textbf{\textcolor{gray}{perm\_view}}, \textbf{\textcolor{gray}{perm\_view\_mod}})},\newline
\newturn
\texttt{struct[used\_in](\textcolor{gray}{Sn}, \textbf{\textcolor{gray}{perm\_view\_mod}})},\newline
\newturn
\texttt{struct[used\_in](\textcolor{gray}{a1}, \textbf{\textcolor{gray}{perm\_view\_mod}})}\\ \hline

(\textbf{---})
In other words, use the fact that for
every subset $M' \subset M$ of size $j$ not
containing $a_{s(1)}+a_{s(2)}+ \ldots +a_{s(j)}$, there is a
way to permute those $j$ numbers to avoid
$M'$. \\
\vspace{-.5\baselineskip}
\newturn
\texttt{perf[\textcolor{perform}{Suggest}](%
                meta[\textcolor{meta}{strategy}](%
                     \textbf{\textcolor{gray}{perm\_view\_mod}},
                     \textbf{\textcolor{gray}{strong\_ind\_n}}))},\newline
\newturn
\texttt{struct[used\_in](\textcolor{gray}{n}, \textbf{\textcolor{gray}{strong\_ind\_n}})},\newline
\newturn
\texttt{struct[used\_in](\textcolor{gray}{n}, \textcolor{gray}{Sn})}\\ \toprule \bottomrule

(\textbf{Haim}) Addressing Dave:
Sorry, indeed I meant: ``Show that for {\bf one}
permutation...'' \\
\vspace{-.5\baselineskip}
\newturn
\texttt{perf[\textcolor{perform}{Retract}](%
                                \textbf{\textcolor{gray}{perm\_view}})}, \newline
\newturn
\texttt{perf[\textcolor{perform}{Agree}](meta[\textcolor{meta}{goal}](%
                                \textbf{\textcolor{gray}{perm\_view\_mod}}))}\\ \toprule
\end{tabular}

\par}

\caption{IATC analysis of MPM1 excerpt (text form)\label{tab:running-example-analysed}}
\end{table}

\FloatBarrier

\begin{figure}
\begin{mdframed}
``\emph{Let $S$ be a finite set of at least two points in the plane. Assume that no three points of $S$ are collinear. A windmill is a process that starts with a line $\ell$ going through a single point $P \in S$. The line rotates clockwise about the pivot $P$ until the first time that the line meets some other point $Q$ belonging to $S$. This point $Q$ takes over as the new pivot, and the line now rotates clockwise about $Q$, until it next meets a point of $S$. This process continues indefinitely. Show that we can choose a point $P$ in $S$ and a line $\ell$ going through $P$ such that the resulting windmill uses each point of $S$ as a pivot infinitely many times.}''

\begin{Verbatim}[commandchars=\\\{\}]
rel[structural](problem, S)
rel[structural](problem, P)
rel[structural](problem, l)
rel[structural](problem, windmill)
rel[structural](problem, pivot)
rel[structural](P, pivot)
rel[structural](P, windmill)
rel[structural](l, windmill)
\end{Verbatim}

1.~\emph{Could you start off with a random point in the plane and prove it doesn't work, if you can't prove that then the opposite holds.}

\begin{Verbatim}[commandchars=\\\{\}]
perf[query](random_test_false)
perf[assert](rel[stronger](rel[not](prove_rtf),
                           rel[not](random_test_false)))
rel[structural](random_test_false, prove_rtf)
\end{Verbatim}

2.~\emph{Connecting the dots: At the point where the pivot changes we create a line that passes through the previous pivot and a new pivot -- like a side of a polygon.}

\begin{Verbatim}[commandchars=\\\{\}]
perf[define](pivot_seq, ps_def)
rel[structural](pivot_seq, pivot)
\end{Verbatim}

2.1.~\emph{Nice. We need only to consider the times when to points are connected -- this gives us a path, and after some time this path will come back to some already visited point. So there is a cycle. If only we could find a cycle which spans all the points, the question is solved\ldots That may be some useful simplification.}

\begin{Verbatim}[commandchars=\\\{\}]
perf[assert](rel[has_property](pivot_seq, has_cycle))
perf[suggest](meta[goal](cycle_spans_S))
perf[judge](value[useful](pivot_seq))
rel[structural](S, cycle_spans_S)
\end{Verbatim}
\end{mdframed}

\caption{IATC tags for the problem and first three comments in MiniPolymath 3. \label{fig:mpm3-example}}
\end{figure}
\paragraph{We observe certain regularities: for example, \texttt{Assert} is present in all five types of comments, but is used most frequently within proof-related comments.}
Annotations from the `\texttt{struct}' grammatical category are most prevalently associated with conjecture-related comments.
(NB.~In this tagging exercise we only considered the `\texttt{used\_in}' facet of the `\texttt{struct}' category, so `\texttt{structural}' is here a synonym for `\texttt{used\_in}'.)
It is not surprising that the performative \texttt{Challenge} is used
most frequently in examples, since, intuitively, an example is likely
to be put forward as a counter-example.  The most prevalent use of
\texttt{Agree} is in comments that are categorised as ``other''.
\texttt{Retract} is frequently used in this category as well, as is
\texttt{stronger} (here, a synonym for `\texttt{implies}').
These usages reflect social
values as well as mathematical semantics.  E.g., one can express support for
an idea by underscoring one's belief in an implication, as in the comment
``Yes, it seems to be a correct solution!'' \cite[\href{https://wp.me/pAG2F-41\#comment-3402}{July 19, 9:35 pm}]{tao2011imo}.

\paragraph{One might suspect that \texttt{Suggest} should be used only within conjectures, but in the current categorisation it is used somewhat more frequently along with concepts.}
This is partly explained by the fact that \texttt{Suggest} can be used
to introduce either a \texttt{goal} or a \texttt{strategy}.  Sometimes goals
represent conceptual tidying, as in ``I guess there is an odd / even
number of point distinction to do'' \cite[\href{https://wp.me/pAG2F-41\#comment-3398}{July 19, 9:31 pm}]{tao2011imo}.

\paragraph{Furthermore, despite of our self-imposed constraint to map each comment only to the most salient of the five categories, in practice a comment may simultaneously introduce a concept along with a conjecture that applies that concept.}
For example the straightforward concept of
``restriction[s] on how the next pivot is chosen''
appears along with the more speculative conjecture  ``Can we
start with a complete graph and all cycles on that graph and just
discard the ones that don't follow the restrictions to converge on the
ones that do?''
\cite[\href{https://wp.me/pAG2F-41\#comment-3368}{July 19, 8:56 pm}]{tao2011imo}.
The need to introduce concepts also applies in the case of more outlandish conjectures,
such as
``It might be fun to use projective duality''
\cite[\href{https://wp.me/pAG2F-41\#comment-3324}{July 19, 8:23 pm}]{tao2011imo}.
However, a concept may suggest a vague method
without raising a conjecture as such,
e.g., ``I'm thinking spirograph rather than convex hull''
\cite[\href{https://wp.me/pAG2F-41\#comment-3349}{July 19, 8:44 pm}]{tao2011imo}.

\paragraph{In sum, the IATC analysis of MiniPolymath 3 shows in detail how individual contributions to the dialogue are comprised.}
In aggregate, this analysis exposes the structural anatomy of a successful collaborative proof.
It should be noted that not all the contributions to MPM3 were equally relevant to the final solution.   By entering the structures in an explicit graphical model in the manner described in Section \ref{iatc-examples:mathoverflow}, graph theoretic analysis could establish, e.g., the centrality of the various concepts used in the content layer, and who introduced them into the conversation.  

\FloatBarrier

\begin{figure}
{\centering
\begin{tabular}{c@{\hspace{-.3in}}c}
\begin{tabular}{l}
\includegraphics[width=.4\textwidth]{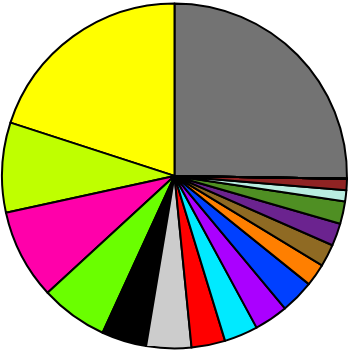}\newline \shfttext{-1in}{Conjecture} \\
\includegraphics[width=.4\textwidth]{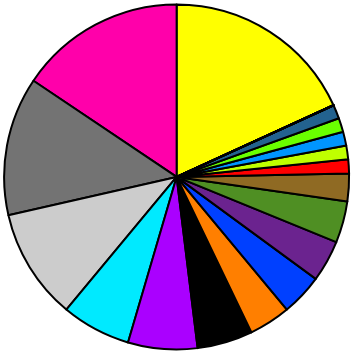}\newline \shfttext{-1in}{Concept} \\
\includegraphics[width=.4\textwidth]{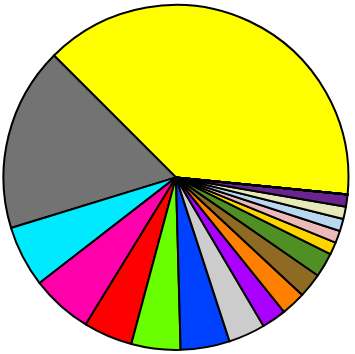}\newline \shfttext{-1.2in}{Proof} \\
\includegraphics[width=.4\textwidth]{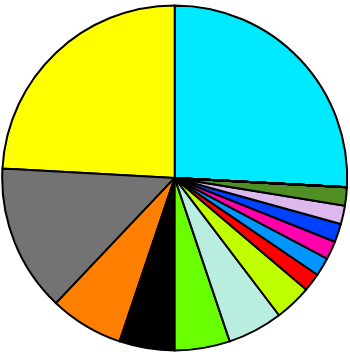}\newline \shfttext{-1in}{Example} \\
\end{tabular}
&
\begin{tabular}{c}
\includegraphics[width=.2\textwidth]{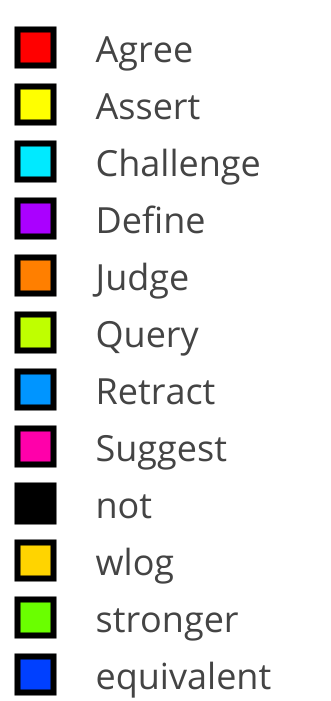}\\[-.11cm]
\shfttext{.12cm}{\includegraphics[width=.2\textwidth]{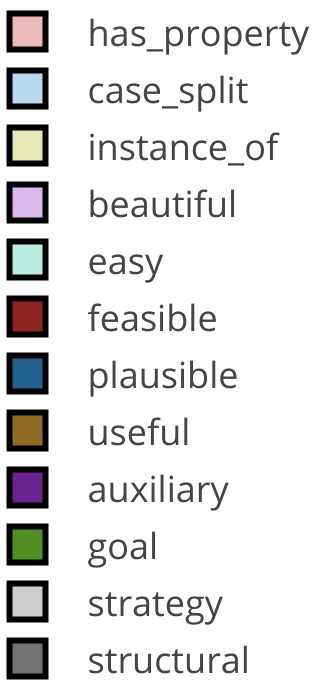}}\\[2cm]
\includegraphics[width=.4\textwidth]{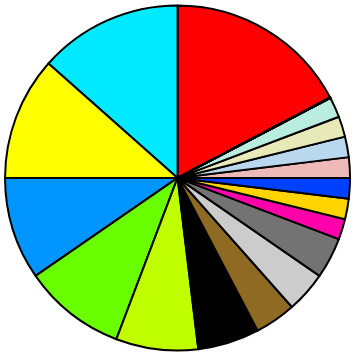} \shfttext{-1in}{Other}
\end{tabular}
\end{tabular}

\par}
\caption{Pie charts showing the relative proportion of IATC tags used to code
  MPM3, across five intuitive kinds of comments.  E.g.,  Comment 1 has been categorised as a \emph{Conjecture}.
The IATC stanza 
\texttt{perf[assert](rel[stronger](rel[not](prove_rtf),
rel[not](random_test_false)))} 
  associated with this comment (see Figure \ref{fig:mpm3-example})
  therefore adds these values to the usage counts
  within the
  \emph{Conjecture} pie chart: `\texttt{Assert}' +1,
  `\texttt{stronger}' +1, and  `\texttt{not}' +2. \label{fig:piecharts}}
\end{figure}

\FloatBarrier
\section{Conclusion}\label{discussion}

\thesis{We have sought to advance the study of mathematical practice from an argumentation-theoretic perspective.}
We introduced Inference Anchoring Theory + Content,
offered a brief comparison with IAT, which it builds upon,
and used three examples to showcase IATC's capabilities.
We showed that:
\begin{itemize}
\item IATC offers a more faithful representation of everyday mathematical
  practice than does, e.g., Lamport-style structured proof.
\item IATC has the potential to support computational reasoning about mathematics by
  surfacing the structural relationships between pieces of
  mathematical content as they appear in discourse.
\item IATC can recover salient elements of discourse within comments, as
  well as the way these contents connect across comments.
\end{itemize}

\paragraph{Some limitations to the approach should be considered when applying the framework.}
We emphasise that these are limitations and not necessarily flaws in the
overall design.  In general, the limitations could be addressed with
extensions to the language.
\begin{itemize}
\item {IATC does not yet handle everything that is said in mathematical dialogues.}
We saw above that IATC nevertheless helps disambiguate the ``other'' category
bracketed by \citet{pease-and-martin}.  
\item {There are places where IATC representations remain bulky, pushing much of the actual reasoning into whatever representation system handles the content layer.}
\item {One related limitation is that implications and assumptions that mathematicians consider ``obvious'' are
typically elided from their discourse, often for valid expository reasons, and that, therefore, unpacking the contextual relationships between statements
typically requires a mathematically trained annotator.}
\item {We introduced a graphical way to segment dialogues, but IATC does not currently have the ability to express context shifts -- although it can compare contexts with `\texttt{analogy}'.}
\end{itemize}
\citet{corneli2018social} survey other relevant frameworks that might
form extensions for a future version of IATC.
More general-purpose formalisms like the W3C's
``PROV'' \cite{groth2013prov} would allow us to say something about
the provenance and evolution of concepts, but would have nothing to
say about the mathematics-specific features that interest us.

In Section \ref{iatc}, we mentioned that Discourse Representation
Theory (DRT) has informed several earlier efforts to model
mathematical discourse.  We are aware of three PhD theses---by Clauss
Zinn \citeyearpar{zinn2004understanding}, Mohan Ganesalingam
\citeyearpar{ganesalingam2013language}, and Marcos Cramer
\citeyearpar{cramer2013proof}---which have made use of somewhat similar
mathematics-specific interpretations of DRT.  Zinn and Cramer
focused on proof checking, and while Ganesalingam looked
at mathematical communication from a linguist's perspective.  However,
he opted to focus exclusively on mathematics in the ``formal mode,''
leaving informal communication about matters such as
``interestingness'' to one side, because they bring with them a
host of additional complications \cite[pp.~7--8]{ganesalingam2013language}.
From a linguistic point of view, DRT is useful in a mathematical setting, in the first instance,
because of its core ability to express ``legitimate antecedents for anaphor''
\cite[p.~50]{ganesalingam2013language}.
In Ganesalingam's work, this basic feature is extended to allow
sidelong references to definite descriptions
(such as `the set of natural numbers')
by ``introducing generalised anaphors which can have presuppositional material
attached to them''
\cite[pp.~25, 237]{ganesalingam2013language}.
Specifically, this allows one to infer from statements such as
``$x$ is prime'' that $x$ is in fact a member of the set of natural numbers (p.~25). 

The associated requirement of combining semantics and pragmatics \cite[p.~336]{doi:10.1093/jos/9.4.333}
is reminiscent of our treatment of unspoken assertions
and unstated features of content in our IATC-based analyses.
To continue the comparison, Ganesalingam's adaptations of DRT
overcame limitations, having to do with quantifier scoping, that constrained
earlier type-theoretic analyses \cite[pp.~81--82]{ganesalingam2013language}.
This is broadly similar to our use of nested structure in Section \ref{iatc-examples:gowers}.
Indeed, \citet{sowa2000knowledge} shows that several different approaches
to nested structure (including DRT)
are all mutually equivalent from a logical point of view.
As indicated by
\citet{doi:10.1093/jos/9.4.333},
pragmatics is relevant for DRT-based models
because it can inform the context-specific resolution of Discourse
Representation Schemes.  This is related to
the question we highlighted in Section \ref{iatc-examples:mathoverflow}:
how to model with the transitions between discourse moves in mathematics?
IAT accounts for similar issues by making reference to dialogue norms,
but we have seen that for mathematical dialogues,
detailed content- and context-specific issues need to be
taken into consideration at each stage.
The models of content evolution used by
\citet{ganesalingam2016fully} to keep track of proof generation were
structurally similar to the DRT-based models developed by
\citet{ganesalingam2013language}: in this case, the evolution was
governed by a limited set of reasoning tactics.  Our work with IATC highlights 
features of mathematical reasoning, like \texttt{analogy}, that more
general heuristics will need to account for.

\paragraph{There are other resources available which could further expand IATC's offerings in this regard.}
For example, a recent special issue of \emph{Argument \& Computation} \cite{ac-special-issue} includes papers detailing the usefulness of
rhetorical structures for argument mining.   \citet{mitrovic2017ontological}, in that volume, indicate the SALT Rhetorical Ontology \cite{groza2012advances} as relevant prior work.  SALT contains three categories---coherence relations, argument scheme relations, and rhetorical blocks---each of which unfolds with considerable further detail.  These three categories can be seen as somewhat analogous to IATC's grammatical categories.   \citet{mitrovic2017ontological} and \citet{lawrence2017harnessing} point to foundational work of \citet{fahnestock1999rhetorical,fahnestock2004figures} on the argumentative function of rhetorical figures, particularly in science writing.  IATC might be profitably connected to such analyses. Furthermore, the integration of rhetoric into argument mining highlights the relevance of structures that are rather different from the IAT-style transitions that have been used in work summarised by \citet{budzynska2015automatically}.  White's \citeyearpar[p.~6]{white1978tropics} pithy assertion that ``logic itself is merely a formalization of tropical strategies'' can serve as an additional provocation to develop structural analyses of this sort.

\paragraph{Nevertheless, whether mathematical content is modelled using ideas from logic, rhetoric, or other sources, considerable further work will be required to effectively describe the processes that are employed in forming and responding to mathematical arguments.}
A small case study included as an appendix to \citet{lak} (and, incidentally, based on MiniPolymath 3) illustrates the plausibility of Lakatos's
model---however that model is clearly far from complete as a theory of mathematical production.
Pease et al were concerned with mathematical content only
insofar as it fills slots for some 20 dialogue moves that are based on
Lakatos's strategy for arguing about lemmas and counterexamples.
For example, $\mathit{MonsterBar}(m, c, r)$
gives a reason $r$, contradicting the justification $m$ for the
counter-conjecture not-$c$.  At no point does this theory touch the
supposed mathematical ground of axioms and rules of inference.  That
the reason $r$, for example, may have been formed inductively, or
deductively, or in some other way, goes undiscussed.  IATC would allow
us to expand the structure that appears within statements like $r$.
Whereas Pease et al's formalisation of Lakatosian reasoning as a
dialogue game offers a computational model of certain dynamical
patterns in mathematics, our current work has focused on kinematics.
The efforts can be seen as complementary: \citet{Bundy20130194} has argued that the right representation can considerably simplify reasoning.

One promising approach to modelling process combines argumentation and multi-agent
systems
\cite{modgil2007towards,maghraby2012automated,Robertson2012}.
However, most approaches to modelling specifically \emph{mathematical}
agents have had significant limitations.
Thus, for example, \citet{fiedler2007argumentation} have described the difficulty of
squaring argumentation-theoretic work with the methods of formal
proof.  Ganesalingam and Gowers's \citeyearpar{ganesalingam2016fully}
project aimed at simulating a solitary individual rather than a
population.  However, \citet{furse1990did} had already called into question the robustness of approaches to modelling mathematical creativity that only model a solitary creative individual.
\citet{pease2009bridging} describe an implementation effort
that made use of a multi-agent approach, drawing on argumentation
theory concepts and a Lakatosian model of dialogue.  However,
the mathematical applications of that system were limited to
straightforward computational aspects of number theory and group theory, which
suggests a ``knowledge bottleneck''
\cite{saint2016argument,moens2017argumentation}.

\paragraph{As indicated in a report of the \citet[p.~90]{national2014developing}, ``knowledge extraction and structuring in the context of mathematics'' is in demand on an increasingly industrial scale.}
IATC allows methods of argumentation to interface with those of
knowledge representation; both aspects are relevant to knowledge
extraction.  Formalisation of IATC would assist in its
applicability: ``IKL Conceptual Graphs'' defined by
\citet{sowa2008conceptual} would provide a natural foundation.  IKL,
the \emph{IKRIS Knowledge Language} \cite{hayes2006ikl,sowa2008conceptual},
deals elegantly with context and has been used as a representational
formalism in a project with aims comparable to our own: the Slate
project \cite{bringsjord2008slate}, which centred on an argumentation
tool that could support a mixture of deductive and informal
reasoning.\footnote{\url{http://www.jfsowa.com/ikl/IKLslate.pdf}}
Previous work on mathematical usage can also inform future efforts in
knowledge modelling with IATC \cite{trzeciak2012mathematical,wells2003handbook,
  wolska2015students,ginev2011structure}.  

\paragraph{Mathematical Knowledge Management, particularly in the ``flexiformal'' understanding developed by \citet{kohlhase2012flexiformalist} and \citet{kohlhase2017mathematical}, presents another paradigm that could eventually be integrated with IATC.}
Flexiformality combines strict formalisations of those parts of
mathematics for which that makes sense with opaque representations of
constants, objects, and informal theories.  \citet{iancu-thesis} built
on Kohlhase's work, and focused on ``co-representing both the
narration and content aspects of mathematical knowledge in a structure
preserving way'' (pp.~3-4).  However, modelling narrative in Iancu's
sense is more relevant to the ``frontstage'' presentation of
mathematics in a single authorial voice than to the ``backstage''
production of mathematics (cf.~\citet{hersh1991mathematics}).  Section
\ref{iatc-examples:mathoverflow} illustrated one such example from
backstage: mathematicians need to be able to choose between different
mathematical problems.

\paragraph{IATC offers a step forward for research into both the communication and production of mathematics, and can play a role in future work on knowledge extraction and simulation.}
Potential applications include, among others, the development of a new
generation of mathematics tutoring software and digital assistants
that engage their users in thought-provoking dialogues.

\section{Acknowledgements}
Our anonymous reviewers offered comments that improved the paper.
The authors also thank Katarzyna Budzynska, Alan Bundy, Pat Hayes, Raymond Puzio, and Chris Reed for helpful discussions, and acknowledge the support of fellow researchers in the ARG-Tech group at the University of Dundee, and the DReaM group at the University of Edinburgh.  Figures were drawn using IHMC CmapTools, the Python libraries matplotlib and plotly, and the TikZ package for \LaTeX.

\bibliographystyle{spbasic}      

\begin{thebibliography}{96}
\providecommand{\natexlab}[1]{#1}
\providecommand{\url}[1]{{#1}}
\providecommand{\urlprefix}{URL }
\expandafter\ifx\csname urlstyle\endcsname\relax
  \providecommand{\doi}[1]{DOI~\discretionary{}{}{}#1}\else
  \providecommand{\doi}{DOI~\discretionary{}{}{}\begingroup
  \urlstyle{rm}\Url}\fi
\providecommand{\eprint}[2][]{\url{#2}}

\bibitem[{Aberdein and Dove(2013)}]{Aberdein2013}
Aberdein A, Dove IJ (2013) Introduction. In: Aberdein A, Dove IJ (eds) The
  Argument of Mathematics, Springer Netherlands, Dordrecht, pp 1--8,
  \urlprefix\url{https://doi.org/10.1007/978-94-007-6534-4_1}

\bibitem[{Aliseda(2003)}]{aliseda2003mathematical}
Aliseda A (2003) Mathematical reasoning vs.~abductive reasoning: a structural
  approach. Synthese 134(1-2):25--44

\bibitem[{Barany(2010)}]{barany2010b}
Barany M (2010) `[{B}]ut this is blog maths and we're free to make up
  conventions as we go along': {P}olymath1 and the modalities of `massively
  collaborative mathematics'. In: Ayers P, Ortega F (eds) Proceedings of the
  6th {I}nternational {S}ymposium on {W}ikis and {O}pen {C}ollaboration, ACM

\bibitem[{Botting(2015)}]{botting2015inferences}
Botting D (2015) Inferences and illocutions. Argument \& Computation
  6(3):246--264

\bibitem[{Bringsjord et~al(2008)Bringsjord, Taylor, Shilliday, Clark, Arkoudas,
  Schoelles, Destefano, and Wodicka}]{bringsjord2008slate}
Bringsjord S, Taylor J, Shilliday A, Clark M, Arkoudas K, Schoelles M,
  Destefano M, Wodicka J (2008) Slate: {An} {Argument-Centered} {Intelligent}
  {Assistant} to {Human} {Reasoners}. In: Grasso F, Green N, Kibble R, Reed C
  (eds) Proceedings of the 8th International Workshop on Computational Models
  of Natural Argument (CMNA 2008), pp 1--10

\bibitem[{Budzynska(2013)}]{budzynska2013circularity}
Budzynska K (2013) Circularity in ethotic structures. Synthese
  190(15):3185--3207

\bibitem[{Budzynska and Reed(2011)}]{budzynska2011speech}
Budzynska K, Reed C (2011) Speech {Acts} of {Argumentation}: {Inference}
  {Anchors} and {Peripheral} {Cues} in {Dialogue}. In: Grasso F, Green N, Reed
  C (eds) Computational Models of Natural Argument: Papers from the 2011 AAAI
  workshop,
  \urlprefix\url{http://www.aaai.org/ocs/index.php/WS/AAAIW11/paper/view/3940}

\bibitem[{Budzynska et~al(2013)Budzynska, Janier, Reed, and
  Saint-Dizier}]{budzynska2013towards}
Budzynska K, Janier M, Reed C, Saint-Dizier P (2013) Towards extraction of
  dialogical arguments. In: Grasso F, Green N, Reed C (eds) Proceedings of 13th
  International Conference on Computational Models of Natural Argument (CMNA
  2013)

\bibitem[{Budzynska et~al(2014{\natexlab{a}})Budzynska, Janier, Kang, Reed,
  Saint-Dizier, Stede, and Yaskorska}]{budzynska2014towards}
Budzynska K, Janier M, Kang J, Reed C, Saint-Dizier P, Stede M, Yaskorska O
  (2014{\natexlab{a}}) Towards argument mining from dialogue. In: Parsons S,
  Oren N, Reed C, Cerutti F (eds) Computational Models of Argument: Proceedings
  of COMMA 2014, IOS Press, Frontiers in Artificial Intelligence and
  Applications, vol 266, pp 185--196,
  \urlprefix\url{http://comma2014.arg.dundee.ac.uk/res/pdfs/19-budzynska.pdf}

\bibitem[{Budzynska et~al(2014{\natexlab{b}})Budzynska, Janier, Reed,
  Saint{-}Dizier, Stede, and Yaskorska}]{budzynska2014model}
Budzynska K, Janier M, Reed C, Saint{-}Dizier P, Stede M, Yaskorska O
  (2014{\natexlab{b}}) A {Model} for {Processing} {Illocutionary} {Structures}
  and {Argumentation} in {Debates}. In: Calzolari N, Choukri K, Declerck T,
  Loftsson H, Maegaard B, Mariani J, Moreno A, Odijk J, Piperidis S (eds)
  Proceedings of the Ninth International Conference on Language Resources and
  Evaluation (LREC-2014), Reykjavik, Iceland, May 26-31, 2014, pp 917--924,
  \urlprefix\url{http://www.lrec-conf.org/proceedings/lrec2014/summaries/77.html}

\bibitem[{Budzynska et~al(2015)Budzynska, Janier, Kang, Konat, Reed,
  Saint-Dizier, Stede, and Yaskorska}]{budzynska2015automatically}
Budzynska K, Janier M, Kang J, Konat B, Reed C, Saint-Dizier P, Stede M,
  Yaskorska O (2015) Automatically identifying transitions between locutions in
  dialogue. In: Mohammed D, Lewinski M (eds) Argumentation and Reasoned Action:
  Proceedings of the 1st European Conference on Argumentation, Lisbon 2015,
  Volume II, College Publications, no.~63 in Studies in Logic and
  Argumentation, pp 311--328

\bibitem[{Budzynska et~al(2016)Budzynska, Janier, Reed, and
  Saint-Dizier}]{budzynska2016theoretical}
Budzynska K, Janier M, Reed C, Saint-Dizier P (2016) Theoretical foundations
  for illocutionary structure parsing. Argument \& Computation 7(1):91--108

\bibitem[{Bundy(1988)}]{bundy1988use}
Bundy A (1988) The use of explicit plans to guide inductive proofs. In: Lusk E,
  Overbeek R (eds) 9th International Conference on Automated Deduction,
  Argonne, Illinois, USA, May 23-26, 1988. Proceedings, Springer, pp 111--120

\bibitem[{Bundy(2013)}]{Bundy20130194}
Bundy A (2013) The interaction of representation and reasoning. Proceedings of
  the Royal Society of London A: Mathematical, Physical and Engineering
  Sciences 469(2157),
  \urlprefix\url{http://rspa.royalsocietypublishing.org/content/469/2157/20130194}

\bibitem[{Carrascal(2015)}]{Carrascal2015}
Carrascal B (2015) Proofs, mathematical practice and argumentation.
  Argumentation 29(3):305--324,
  \urlprefix\url{http://dx.doi.org/10.1007/s10503-014-9344-0}

\bibitem[{{Chandrasekhar} et~al(2010)}]{chandrasekhar2010group}
{Chandrasekhar}, et~al (2010) Group cannot be the union of conjugates.
  \urlprefix\url{http://mathoverflow.net/q/34044}

\bibitem[{Corneli et~al(2017{\natexlab{a}})Corneli, Martin, Murray-Rust, and
  Pease}]{corneli2017towards}
Corneli J, Martin U, Murray-Rust D, Pease A (2017{\natexlab{a}}) Towards
  mathematical {AI} via a model of the content and process of mathematical
  question and answer dialogues. In: Geuvers H, England M, Hasan O, Rabe F,
  Teschke O (eds) Intelligent Computer Mathematics 10th International
  Conference, CICM 2017, Edinburgh, UK, 2017, Proceedings

\bibitem[{Corneli et~al(2017{\natexlab{b}})Corneli, Martin, Murray-Rust, Pease,
  Puzio, and Rino~Nesin}]{corneli2017modelling}
Corneli J, Martin U, Murray-Rust D, Pease A, Puzio R, Rino~Nesin G
  (2017{\natexlab{b}}) Modelling the way mathematics is actually done. In:
  Sperber M, Bresson J, Santolucito M, McLean A (eds) 2017 International
  Workshop on Functional Art, Music, Modelling and Design (FARM 2017), ACM

\bibitem[{Corneli et~al(2018)Corneli, Pease, and Stefanou}]{corneli2018social}
Corneli J, Pease A, Stefanou D (2018) Social {Aspects} of {Concept}
  {Invention}. In: Confalonieri R, Pease A, Schorlemmer M, Besold T, Kutz O,
  Maclean E, Kaliakatsos-Papakostas M (eds) Concept Invention: Foundations,
  Implementation, Social Aspects and Applications, Computational Synthesis and
  Creative Systems, Springer

\bibitem[{Cramer(2013)}]{cramer2013proof}
Cramer M (2013) Proof-checking mathematical texts in controlled natural
  language. PhD thesis, Mathematisch-Naturwissenschaftlichen Fakult\"at der
  Rheinischen Friedrich-Wilhelms-Universit\"at Bonn

\bibitem[{Dauphin and Cramer(2018)}]{dauphin2018aspicend}
Dauphin J, Cramer M (2018) {ASPIC}-{END}: {Structured} {Argumentation} with
  {Explanations} and {Natural} {Deduction}. In: Black E, Modgil S, Oren N (eds)
  Theory and Applications of Formal Argumentation 4th International Workshop,
  TAFA 2017, Melbourne, VIC, Australia, August 19-20, 2017, Revised Selected
  Papers, Springer, no. 10757 in LNAI, pp 51--66

\bibitem[{Dove(2009)}]{dove2009towards}
Dove IJ (2009) Towards a theory of mathematical argument. Foundations of
  Science 14(1-2):137--152

\bibitem[{Dufour(2013)}]{Dufour2013}
Dufour M (2013) Arguing Around Mathematical Proofs, Springer Netherlands,
  Dordrecht, pp 61--76.
  \urlprefix\url{https://doi.org/10.1007/978-94-007-6534-4_5}

\bibitem[{Dutilh~Novaes(2016)}]{novaes2016reductio}
Dutilh~Novaes C (2016) Reductio ad absurdum from a dialogical perspective.
  Philosophical Studies 173(10):2605--2628

\bibitem[{Epstein(2015)}]{epstein2015wanted}
Epstein SL (2015) Wanted: {Collaborative} intelligence. Artificial Intelligence
  221:36--45

\bibitem[{Fahnestock(1999)}]{fahnestock1999rhetorical}
Fahnestock J (1999) Rhetorical {Figures} in {Science}. Oxford University Press

\bibitem[{Fahnestock(2004)}]{fahnestock2004figures}
Fahnestock J (2004) Figures of {Argument} ({OSSA} 2005 {Keynote} {Address}).
  Informal Logic 24(2)

\bibitem[{Fiedler and Horacek(2007)}]{fiedler2007argumentation}
Fiedler A, Horacek H (2007) Argumentation within deductive reasoning.
  International Journal of Intelligent Systems 22(1):49--70

\bibitem[{Furse(1990)}]{furse1990did}
Furse E (1990) Why did {AM} run out of steam? Tech. rep., CS-90-4, Department
  of Computer Studies, University of Glamorgan

\bibitem[{Ganesalingam(2013)}]{ganesalingam2013language}
Ganesalingam M (2013) The {Language} of {Mathematics}, {A} {Linguistic} and
  {Philosophical} {Investigation}, LNCS, vol 7805. Springer Verlag

\bibitem[{Ganesalingam and Gowers(2017)}]{ganesalingam2016fully}
Ganesalingam M, Gowers W (2017) A {Fully} {Automatic} {Theorem} {Prover} with
  {Human}-{Style} {Output}. Journal of Automated Reasoning 58:253--291

\bibitem[{Gasteren(1990)}]{gasteren1990shape}
Gasteren AJ (1990) On the shape of mathematical arguments, Lecture Notes in
  Computer Science, vol 445. Springer Science \& Business Media

\bibitem[{Ginev(2011)}]{ginev2011structure}
Ginev D (2011) The structure of mathematical expressions. Master's thesis,
  Jacobs University, Bremen, Germany

\bibitem[{Gowers(2017)}]{gowers-talk-ini}
Gowers W (2017) How do human mathematicians avoid big searches?
  \urlprefix\url{https://www.newton.ac.uk/seminar/20170728133014301},
  {Lecture}, Fri, July 28, 13:30 -- 14:30, Isaac Newton Institute, University
  of Cambridge

\bibitem[{Gowers and Ganesalingam(2012)}]{gowers-talk}
Gowers W, Ganesalingam M (2012) Modelling the mathematical discovery process.
  Maxwell Institute Lecture, Fri, November 2, 4pm -- 5pm, James Clerk Maxwell
  Building, University of Edinburgh

\bibitem[{Groth and Moreau(2013)}]{groth2013prov}
Groth P, Moreau L (2013) {PROV-Overview}. {An} {Overview} of the {PROV}
  {Family} of {Documents}. World Wide Web Consortium,
  \urlprefix\url{https://www.w3.org/TR/prov-overview/}

\bibitem[{Groza(2012)}]{groza2012advances}
Groza T (2012) Advances in semantic authoring and publishing, Studies on the
  Semantic Web, vol~13. IOS Press

\bibitem[{Harris and Marco(2017)}]{ac-special-issue}
Harris R, Marco CD (2017) Argument \& Computation 8(3) [Special Issue on
  Rhetorical figures, arguments, computation.]

\bibitem[{Hayes(2006)}]{hayes2006ikl}
Hayes P (2006) {IKL} guide.
  \urlprefix\url{http://www.ihmc.us/users/phayes/IKL/GUIDE/GUIDE.html}

\bibitem[{Hendrix(1975)}]{hendrix1975partitioned}
Hendrix GG (1975) Partitioned networks for the mathematical modeling of natural
  language semantics. PhD thesis, The University of Texas at Austin, also
  archived as Tech. Report NL-28, Department of Computer Science

\bibitem[{Hendrix(1979)}]{hendrix1979encoding}
Hendrix GG (1979) Encoding knowledge in partitioned networks. In: Findler NV
  (ed) Associative networks: Representation and use of knowledge by computers,
  Academic Press, pp 51--92

\bibitem[{Hersh(1991)}]{hersh1991mathematics}
Hersh R (1991) Mathematics has a front and a back. Synthese 88(2):127--133

\bibitem[{Iancu(2017)}]{iancu-thesis}
Iancu M (2017) Towards {Flexiformal} {Mathematics}. PhD thesis, Jacobs
  University, Bremen, DE

\bibitem[{Inglis et~al(2007)Inglis, Mejia-Ramos, and
  Simpson}]{inglis2007modelling}
Inglis M, Mejia-Ramos JP, Simpson A (2007) Modelling mathematical
  argumentation: The importance of qualification. Educational Studies in
  Mathematics 66(1):3--21

\bibitem[{Janier and Reed(2017)}]{janier2017towards}
Janier M, Reed C (2017) Towards a theory of close analysis for dispute
  mediation discourse. Argumentation 31(1):45--82

\bibitem[{Kamp and Reyle(1993)}]{kamp1993discourse}
Kamp H, Reyle U (1993) {From} {Discourse} {to} {Logic} {Introduction} {to}
  {Modeltheoretic} {Semantics} {of} {Natural} {Language}, {Formal} {Logic}
  {and} {Discourse} {Representation} {Theory}, Studies in Linguistics and
  Philosophy, vol~42. Springer-Science+Business Media

\bibitem[{Karttunen(1976)}]{karttunen1976discourse}
Karttunen L (1976) Discourse referents. In: McCawley JD (ed) Syntax and
  Semantics, Vol. 7: Notes from the Linguistic Underground, Academic Press, pp
  363--386

\bibitem[{Klaus(2004)}]{klaus2004content}
Klaus K (2004) Content analysis: An introduction to its methodology, 2nd edn.
  Sage Publications

\bibitem[{Kohlhase(2012)}]{kohlhase2012flexiformalist}
Kohlhase M (2012) The flexiformalist manifesto. In: Voronkov A, Negru V, Ida T,
  Jebelean T, Petcu D, Watt S, Zaharie D (eds) Symbolic and Numeric Algorithms
  for Scientific Computing (SYNASC 2012), 14th International Symposium on,
  IEEE, pp 30--35

\bibitem[{Kohlhase et~al(2017)Kohlhase, Koprucki, M{\"u}ller, and
  Tabelow}]{kohlhase2017mathematical}
Kohlhase M, Koprucki T, M{\"u}ller D, Tabelow K (2017) Mathematical models as
  research data via flexiformal theory graphs. In: Geuvers H, England M, Hasan
  O, Rabe F, Teschke O (eds) Intelligent Computer Mathematics 10th
  International Conference, CICM 2017, Edinburgh, UK, July 17-21, 2017,
  Proceedings, Springer, LNAI, vol 10383, pp 224--238

\bibitem[{Lakatos(1976)}]{lakatos2015proofs}
Lakatos I (1976) Proofs and refutations: The logic of mathematical discovery.
  Cambridge University Press

\bibitem[{Lamport(1995)}]{lamport1995write}
Lamport L (1995) How to write a proof. The {A}merican {M}athematical {M}onthly
  102(7):600--608

\bibitem[{Lamport(1999)}]{lamport1999specifying}
Lamport L (1999) Specifying {Concurrent} {Systems} with {TLA}{$^{+}$}. NATO
  Science Series, III: Computer and Systems Sciences 173(173):183--247

\bibitem[{Lamport(2012)}]{lamport2012write21st}
Lamport L (2012) How to write a 21st century proof. Journal of fixed point
  theory and applications 11(1):43--63

\bibitem[{Lamport(2014)}]{lamport2014tla2}
Lamport L (2014) {TLA}$^{+2}$: {A} {Preliminary} {Guide}.
  \urlprefix\url{http://lamport.azurewebsites.net/tla/tla2-guide.pdf}

\bibitem[{Lamport(2015)}]{lamport2015tlahyperbook}
Lamport L (2015) The {TLA}$^{+}$ hyperbook.
  \urlprefix\url{http://lamport.azurewebsites.net/tla/hyperbook.html}

\bibitem[{Larvor(2012)}]{larvor2012think}
Larvor B (2012) How to think about informal proofs. Synthese 187(2):715--730

\bibitem[{Lawrence et~al(2012)Lawrence, Bex, Reed, and
  Snaith}]{lawrence2012aifdb}
Lawrence J, Bex F, Reed C, Snaith M (2012) {AIF}db: Infrastructure for the
  argument web. In: Verheij B, Szeider S, Woltran S (eds) Computational Models
  of Argument: Proceedings of COMMA 2012, IOS Press, Frontiers in Artificial
  Intelligence and Applications, vol 245, pp 515--516

\bibitem[{Lawrence et~al(2017)Lawrence, Visser, and
  Reed}]{lawrence2017harnessing}
Lawrence J, Visser J, Reed C (2017) Harnessing rhetorical figures for argument
  mining. Argument \& Computation 8(3):289--310

\bibitem[{Maghraby et~al(2012)Maghraby, Robertson, Grando, and
  Rovatsos}]{maghraby2012automated}
Maghraby A, Robertson D, Grando A, Rovatsos M (2012) Automated deployment of
  argumentation protocols. In: Verheij B, Szeider S, Woltran S (eds)
  Computational Models of Argument: Proceedings of COMMA 2012, IOS Press,
  Frontiers in Artificial Intelligence and Applications, vol 245, pp 197--204

\bibitem[{Martin(2015)}]{Martin2015}
Martin U (2015) Stumbling Around in the Dark: Lessons from Everyday
  Mathematics, Springer International Publishing, Cham, pp 29--51.
  \urlprefix\url{https://doi.org/10.1007/978-3-319-21401-6_2}

\bibitem[{Martin and Pease(2013)}]{martin2013does}
Martin U, Pease A (2013) What does mathoverflow tell us about the production of
  mathematics? In: Novak J, Jaimes A (eds) SOHUMAN, 2nd International Workshop
  on Social Media for Crowdsourcing and Human Computation, at ACM Web Science
  2013, May 1, 2013, Paris, \urlprefix\url{https://arxiv.org/abs/1305.0904}

\bibitem[{Mercier and Sperber(2011)}]{mercier2011humans}
Mercier H, Sperber D (2011) Why do humans reason? {Arguments} for an
  argumentative theory. Behavioral and brain sciences 34(2):57--74

\bibitem[{Mitrovi{\'c} et~al(2017)Mitrovi{\'c}, O’Reilly, Mladenovi{\'c}, and
  Handschuh}]{mitrovic2017ontological}
Mitrovi{\'c} J, O’Reilly C, Mladenovi{\'c} M, Handschuh S (2017) Ontological
  representations of rhetorical figures for argument mining. Argument \&
  Computation 8(3):267--287

\bibitem[{Modgil and McGinnis(2007)}]{modgil2007towards}
Modgil S, McGinnis J (2007) Towards {Characterising} {Argumentation} {Based}
  {Dialogue} in the {Argument} {Interchange} {Format}. In: Rahwan I, Parsons S,
  Reed C (eds) Argumentation in Multi-Agent Systems: 4th International
  Workshop, ArgMAS 2007, Honolulu, HI, USA, May 15, 2007, Revised Selected and
  Invited Papers, Springer, pp 80--93

\bibitem[{Moens(2018)}]{moens2017argumentation}
Moens MF (2018) Argumentation mining: {How} can a machine acquire common sense
  and world knowledge? Argument \& Computation 9(1)

\bibitem[{{National Research Council}(2014)}]{national2014developing}
{National Research Council} (2014) Developing a 21st {Century} {Global}
  {Library} for {Mathematics} {Research}. National Academies Press

\bibitem[{Nielsen et~al(2009--2018)}]{polymath-wiki}
Nielsen M, et~al (2009--2018) Polymath wiki.
  \urlprefix\url{http://michaelnielsen.org/polymath1/index.php?title=Main_Page}

\bibitem[{van Oers(2002)}]{vanOers}
van Oers B (2002) Fruits of {Polyphony}: {A} {Commentary} on a
  {Multiperspective} {Analysis} of {Mathematical} {Discourse}. Journal of the
  Learning Sciences 11(2-3):359--363,
  \urlprefix\url{https://doi.org/10.1080/10508406.2002.9672143}

\bibitem[{Pease and Martin(2012)}]{pease-and-martin}
Pease A, Martin U (2012) Seventy four minutes of mathematics: {An} analysis of
  the third {Mini}-{Polymath} project. In: Larvor B, Pease A (eds) Proceedings
  of {AISB}/{IACAP} 2012, {Symposium} on {Mathematical} {Practice} and
  {Cognition} {II},
  \urlprefix\url{http://homepages.inf.ed.ac.uk/apease/papers/seventy-four.pdf}

\bibitem[{Pease et~al(2009)Pease, Smaill, Colton, and Lee}]{pease2009bridging}
Pease A, Smaill A, Colton S, Lee J (2009) Bridging the gap between
  argumentation theory and the philosophy of mathematics. Foundations of
  Science 14(1):111--135

\bibitem[{Pease et~al(2017)Pease, Lawrence, Budzynska, Corneli, and Reed}]{lak}
Pease A, Lawrence J, Budzynska K, Corneli J, Reed C (2017) Lakatos-style
  {Collaborative} {Mathematics} through {Dialectical}, {Structured} and
  {Abstract} {Argumentation}. Artificial Intelligence 246:181--219,
  \urlprefix\url{http://www.sciencedirect.com/science/article/pii/S0004370217300267}

\bibitem[{Pedemonte(2007)}]{pedemonte2007}
Pedemonte B (2007) How can the relationship between argumentation and proof be
  analysed? Educational Studies in Mathematics 66(1):23--41,
  \urlprefix\url{http://dx.doi.org/10.1007/s10649-006-9057-x}

\bibitem[{Reed and Budzynska(2010)}]{reed2011dialogues}
Reed C, Budzynska K (2010) How dialogues create arguments. In: van Eemeren F,
  Garssen B, Godden D, Mitchell G (eds) Proceedings of the 7th Conference of
  the International Society for the Study of Argumentation (ISSA 2010)

\bibitem[{Reed et~al(2017)Reed, Budzynska, Duthie, Janier, Konat, Lawrence,
  Pease, and Snaith}]{Reed2017}
Reed C, Budzynska K, Duthie R, Janier M, Konat B, Lawrence J, Pease A, Snaith M
  (2017) The argument web: an online ecosystem of tools, systems and services
  for argumentation. Philosophy {\&} Technology 30(2):137--160,
  \urlprefix\url{http://dx.doi.org/10.1007/s13347-017-0260-8}

\bibitem[{Robertson(2012)}]{Robertson2012}
Robertson D (2012) Lightweight Coordination Calculus for Agent Systems:
  Retrospective and Prospective, Springer Berlin Heidelberg, Berlin,
  Heidelberg, pp 84--89.
  \urlprefix\url{https://doi.org/10.1007/978-3-642-29113-5_7}

\bibitem[{Robinson(1965)}]{robinson1965machine}
Robinson JA (1965) A machine-oriented logic based on the resolution principle.
  Journal of the ACM (JACM) 12(1):23--41

\bibitem[{Saint-Dizier(2016)}]{saint2016argument}
Saint-Dizier P (2016) Argument {Mining}: The bottleneck of knowledge and
  language resources. In: Calzolari N, Choukri K, Declerck T, Goggi S,
  Grobelnik M, Maegaard B, Mariani J, Mazo H, Moreno A, Odijk J, Piperidis S
  (eds) 10th International Conference on Language Resources and Evaluation
  (LREC 2016), pp 983--990

\bibitem[{van~der Sandt(1992)}]{doi:10.1093/jos/9.4.333}
van~der Sandt RA (1992) Presupposition {Projection} as {Anaphora} {Resolution}.
  Journal of Semantics 9(4):333--377,
  \urlprefix\url{http://dx.doi.org/10.1093/jos/9.4.333}

\bibitem[{Snaith and Reed(2012)}]{Snaith2012}
Snaith M, Reed C (2012) {TOAST}: online {ASPIC}+ implementation. In: Verheij B,
  Szeider S, Woltran S (eds) Computational Models of Argument: Proceedings of
  COMMA 2012, IOS Press, Frontiers in Artificial Intelligence and Applications,
  vol 245, pp 509--510

\bibitem[{Snaith and Reed(2016)}]{snaith2016dialogue}
Snaith M, Reed C (2016) Dialogue grammar induction. In: Mohammed D, Lewinski M
  (eds) Argumentation and Reasoned Action, Volume I: Proceedings of the 1st
  European Conference on Argumentation, Lisbon 2015, Studies in Logic and
  Argumentation, vol~62, College Publications

\bibitem[{Snaith et~al(2010)Snaith, Devereux, Lawrence, and
  Reed}]{snaith2010pipelining}
Snaith M, Devereux J, Lawrence J, Reed C (2010) Pipelining argumentation
  technologies. In: Baroni P, Cerutti F, Giacomin M, R SG (eds) Computational
  Models of Argument: Proceedings of COMMA 2010, IOS Press, Frontiers in
  Artificial Intelligence and Applications, vol 216, pp 447--454

\bibitem[{Sowa(2000)}]{sowa2000knowledge}
Sowa JF (2000) Knowledge representation: logical, philosophical, and
  computational foundations. MIT Press

\bibitem[{Sowa(2008)}]{sowa2008conceptual}
Sowa JF (2008) Conceptual {Graphs}. In: van Harmelen F, Lifschitz V, Porter B
  (eds) Handbook of Knowledge Representation, Elsevier, chap~5, pp 213--237

\bibitem[{Sowa and Majumdar(2003)}]{sowa2003analogical}
Sowa JF, Majumdar AK (2003) Analogical reasoning. In: Aldo A, Lex W, Ganter B
  (eds) Conceptual Structures for Knowledge Creation and Communication: 11th
  International Conference on Conceptual Structures, ICCS 2003, Dresden,
  Germany, July 21-25, 2003, Proceedings, Springer, no. 2746 in LNAI, pp 16--36

\bibitem[{Tanswell(2015)}]{tanswell2015problem}
Tanswell F (2015) A {Problem} with the {Dependence} of {Informal} {Proofs} on
  {Formal} {Proofs}. Philosophia Mathematica 23(3):295,
  \urlprefix\url{http://dx.doi.org/10.1093/philmat/nkv008}

\bibitem[{Tao et~al(2009)}]{tao2009imo}
Tao T, et~al (2009) {IMO} 2009 {Q6} as a mini-polymath project.
  \urlprefix\url{https://wp.me/p3qzP-Ef}

\bibitem[{Tao et~al(2011)}]{tao2011imo}
Tao T, et~al (2011) Minipolymath3 project: 2011 {IMO}.
  \urlprefix\url{https://wp.me/pAG2F-41}

\bibitem[{Trzeciak(2012)}]{trzeciak2012mathematical}
Trzeciak J (2012) Mathematical {English} {Usage}. {A} {Dictionary}.
  \urlprefix\url{http://www.emis.de/monographs/Trzeciak/biglist.html}

\bibitem[{Visser et~al(2011)Visser, Bex, Reed, and
  Garssen}]{visser2011correspondence}
Visser J, Bex F, Reed C, Garssen B (2011) Correspondence between the
  pragma-dialectical discussion model and the {Argument} {Interchange}
  {Format}. Studies in Logic, Grammar and Rhetoric 23(36):189--224

\bibitem[{Walton et~al(2008)Walton, Reed, and Macagno}]{walton08}
Walton D, Reed C, Macagno F (2008) Argumentation Schemes. Cambridge University
  Press, New York, USA

\bibitem[{Wells(2003)}]{wells2003handbook}
Wells C (2003) A handbook of mathematical discourse. Infinity Publishing

\bibitem[{White(1978)}]{white1978tropics}
White HV (1978) Tropics of discourse: essays in cultural criticism. Johns
  Hopkins University Press

\bibitem[{Wolska(2015)}]{wolska2015students}
Wolska MA (2015) Students{'} language in computer-assisted tutoring of
  mathematical proofs. No.~40 in Saarbr{\"u}cken Dissertations in Language
  Science and Technology, Universaar

\bibitem[{Zack and Graves(2001)}]{Zack2001}
Zack V, Graves B (2001) Making mathematical meaning through dialogue: {``}once
  you think of it, the {Z} minus three seems pretty weird{''}. Educational
  Studies in Mathematics 46(1):229,
  \urlprefix\url{https://doi.org/10.1023/A:1014045408753}

\bibitem[{Zinn(2004)}]{zinn2004understanding}
Zinn C (2004) Understanding informal mathematical discourse. PhD thesis,
  Institut fur Informatik, Universitat Erlangen-Nurnberg

\end{thebibliography}

\clearpage
\appendix
\section{Reference coding samples} \label{app:reference-coding-samples}
This appendix collects sample texts and IATC codings to
supplement Tables \ref{iatc-table:1} and \ref{iatc-table:2} in Section
\ref{iatc}, which introduced the available codes.  Texts are sourced from the
examples discussed in Section \ref{iatc-examples}.    In general, one utterance may
expand to multiple statements in IATC; accordingly, texts may
appear here multiple times.  Bold face is used to illustrate the
portion of the text, at right, that justifies the tag that appears, at
left.   Numbering refers
to the tree-ordering of MiniPolymath 3 comments, unless another source
is indicated.

\medskip

{\centering
\textbf{Performatives} 

\par}
\medskip
\noindent\begin{tabular}{|p{.45\textwidth}|p{.45\textwidth}|}
\hline
perf[\textbf{assert}](rel[implies](rel[not](prove_rtf), rel[not](random_test_false))) & 1.~Could you start off with a random point in the plane and prove it doesn't work, \textbf{if you can't prove that then the opposite holds.}\\ \hline
perf[\textbf{agree}](cycle_partition) & 2.2.1.1.~\textbf{I believe this is true.} It proves that it's enough to find a cycle that visits each vertex at least once.\\ \hline
perf[\textbf{challenge}](problem, equi_tri_stuck) & 3.1.1.~Say there are four points: an equilateral triangle, and then one point in the center of the triangle. No three points are collinear. \textbf{It seems to me that the windmill can not use the center point more than once! As soon as it hits one of the corner points, it will cycle indefinitely through the corners and never return to the center point.} I must be missing something here.\\ \hline
perf[\textbf{retract}](perf[challenge](problem, equi_tri_stuck)) & 3.1.1.2.1.~\textbf{Ohhh\ldots I misunderstood the problem. I saw it as a half-line extending out from the last point, in which case you would get stuck on the convex hull.} But apparently it means a full line, so that the next point can be ``behind'' the previous point. \textbf{Got it.} \\ \hline
perf[\textbf{define}](pivot_seq, ps_def) & 2.~Connecting the dots: \textbf{At the point where the pivot changes we create a line that passes through the previous pivot and a new pivot -- like a side of a polygon}.\\ \hline
perf[\textbf{suggest}](meta[goal](cycle_spans_S)) & 2.1.~Nice. We need only to consider the times when two points are connected -- this gives us a path, and after some time this path will come back to some already visited point. So there is a cycle. \textbf{If only we could find a cycle which spans all the points, the question is solved}. That may be some useful simplification.\\ \hline
perf[\textbf{judge}](value[useful](pivot_seq)) & 2.1.~Nice. We need only to consider the times when two points are connected -- this gives us a path, and after some time this path will come back to some already visited point. So there is a cycle. If only we could find a cycle which spans all the points, the question is solved. \textbf{That may be some useful simplification.} \\ \hline
perf[\textbf{query}](random_test_false) & 1.~\textbf{Could you start off with a random point in the plane and prove it doesn't work}, if you can't prove that then the opposite holds. \\ \hline
perf[\textbf{queryE}](additional_condition_on cycles(X)) & 2.1.1.1.~For example, the restriction on how the next pivot is chosen (geometrically: comment 9). \textbf{Are there any other restrictions?} Can we start with a complete graph and all cycles on that graph and just discard the ones that don't follow the restrictions to converge on the ones that do?\\ \hline
\end{tabular}

\newpage
{\centering
\textbf{Inferential Structure}  

\par}

\medskip

\noindent
\noindent\begin{tabular}{|p{.45\textwidth}|p{.45\textwidth}|}
\hline
perf[assert](rel[\textbf{implies}](rel[not](prove_rtf), rel[not](random_test_false))) & 1.~Could you start off with a random point in the plane and prove it doesn't work, \textbf{if you can't prove that then the opposite holds.}\\ \hline
perf[assert](rel[\textbf{equivalent}](problem, forall_exists_problem), cycle_partition) & 2.2.1.1.~I believe this is true. It proves that \textbf{it's enough to find a cycle that visits each vertex at least once}. There are no ``rho'' processes with an initial segment that doesn't repeat.\\ \hline
perf[assert](rel[implies](rel[not](prove_rtf), rel[\textbf{not}](random_test_false))) & 1.~Could you start off with a random point in the plane and \textbf{prove it doesn't work}, if you can't prove that then the opposite holds.\\ \hline
perf[question](rel[implies](rel[\textbf{conjunction}](G_infinite_group, H_subgroup_of_G, H_finite_index_in_G), G_not_equal_to_union_of_cosets)) & (Section \ref{iatc-examples:mathoverflow}, Question)~I have seen this problem, that if \textbf{$G$
is a finite group and $H$ is a proper subgroup of $G$ with finite index} then $ G \neq \cup_{g \in G} gHg^{-1}$. Does this remain true for the infinite case also? \\ \hline
 perf[assert](rel[\textbf{has_property}](pivot_seq, has_cycle)) & 2.1.~Nice. We need only to consider the times when two points are connected -- \textbf{this gives us a path, and after some time this path will come back to some already visited point. So there is a cycle.} If only we could find a cycle which spans all the points, the question is solved. That may be some useful simplification.\\ \hline
rel[\textbf{instance_of}](S, convex_plus_point) & 3.1.~Yes. Can we \textbf{do it if there is a single point not on the convex hull of the points?}\\ \hline
perf[assert]( rel[\textbf{indep_of}]( disj_path, M )) & (MPM1, 31.)~Quick thought following on David Speyer's first comment:  The problem asks us to prove that no set of size $(n-1)$ can disconnect two diagonally opposing vertices in the $n$-cube.  By Menger's theorem, this is equivalent to proving that there are n internally vertex-disjoint paths between these two vertices.  So, now we are faced with \textbf{a constructive problem, independent of the set $M$:  Construct $n$ vertex-disjoint paths from $0^n$ to $1^n$ in the $n$-cube.}\\ \hline
perf[assert](rel[\textbf{case_split}](IS, IS_A, IS_B)) & 2.1.1.1.1.~The line must sweep out a full rotation (and only one full rotation) of $2\pi$ during the traversal of $S$. I feel like this is intimately related to proving that there is a starting angle for any point $P$ in $S$ such that all of $S$ is then traversed. I'm trying to show this by induction. Base case ($|S|=2$) is obvious. \textbf{Let $|S| = n$, take $S' = S \cup \{Q\}$, and start with some windmill traversal of $S$.} \textbf{Case A:} $Q$ is unreachable. Therefore we just traverse $S$, taking $2\pi$ to do so by induction. \textbf{Case B:} $Q$ is reachable at some angle. [\ldots] \\ \hline
perf[assert](rel[\textbf{wlog}](problem, zero_angle), one_turn) & 11.2.3.1.~Only the starting point matters. By the problem statement, it appears that the initial angle is irrelevant to the existence of a pivot point $P^*$ from which all of $S$ is traversed. Every point in S is a pivot point, but only with a specific range of starting angle (e.g. those consistent with the cycle generating $S$). The union of these intervals must necessarily be $[0,2\pi)$, and thus \textbf{we can assume WLOG that the starting angle is 0} (and thus we single out a specific point -- or points in the case of $|S| = 2$).\\
\hline
\end{tabular}
\newpage

{\centering
\textbf{Heuristic Value Judgments} 

\par}

\smallskip

\noindent
\noindent\begin{tabular}{|p{.45\textwidth}|p{.45\textwidth}|}
\hline
perf[judge](value[\textbf{easy}](S_is_conv)) & 3.~\textbf{If the points form a convex polygon, it is easy.}\\ \hline
perf[judge](rel[not](value[\textbf{plausible}](equi_tri_stuck))) & 3.1.1.~Say there are four points: an equilateral triangle, and then one point in the center of the triangle. No three points are collinear. It seems to me that the windmill can not use the center point more than once! As soon as it hits one of the corner points, it will cycle indefinitely through the corners and never return to the center point. \textbf{I must be missing something here.}\\ \hline
perf[judge](value[\textbf{beautiful}](proof_sugg)) & 14.2.1.~\textbf{Very nice!} Don't we run into problems with a convex hull though? Take a square with a point in the middle ($M$) and pass the diagonal of the square (not through $M$) -- it seems to me M is never visited (though I may be wrong here). I think we should be more specific in our initial choice of line, maybe?\\ \hline
perf[judge](value[\textbf{useful}](pivot_seq)) & 2.1.~Nice. We need only to consider the times when two points are connected -- this gives us a path, and after some time this path will come back to some already visited point. So there is a cycle.  \textbf{If only we could find a cycle which spans all the points, the question is solved. That may be some useful simplification.}\\ \hline
\end{tabular}

\medskip

{\centering
\textbf{Reasoning Tactics} 

\par}

\smallskip

\noindent
\noindent\begin{tabular}{|p{.45\textwidth}|p{.45\textwidth}|}
\hline
perf[suggest](meta[\textbf{goal}](cycle_spans_S)) & 2.1.~Nice. We need only to consider the times when two points are connected -- this gives us a path, and after some time this path will come back to some already visited point. So there is a cycle. \textbf{If only we could find a cycle which spans all the points, the question is solved}. That may be some useful simplification.\\ \hline
perf[suggest](meta[\textbf{strategy}](cycle_spans_S, process_of_elim)) & 2.1.1.1.~For example, the restriction on how the next pivot is chosen (geometrically: comment 9). Are there any other restrictions? \textbf{Can we start with a complete graph and all cycles on that graph and just discard the ones that don't follow the restrictions to converge on the ones that do?}\\ \hline
perf[suggest](meta[\textbf{auxiliary}](problem, forall_split)) & Ok. I think the solution might involve this observation, \textbf{with the observation that every point participates in a ``splitting'' line (one with $n/2$ points on one side).}\\ \hline
perf[assert](meta[\textbf{analogy}](compute 500th digit of (sqrt(2)+sqrt(3))\textasciicircum2012, compute 500th digit of (x+y)\textasciicircum2012)) & (Section \ref{iatc-examples:gowers}) \textbf{Can we do this for $x+y$?} For $e$?  Rationals with small denominator?\\ \hline
perf[assert](rel[\textbf{implements}](\#SUBGRAPH, the trick might be))  & (Section \ref{iatc-examples:gowers}) \textbf{$(\sqrt{2}+\sqrt{3})^{2012}+(\sqrt{3}-\sqrt{2})^{2012}$ is an integer! And $(\sqrt{3}-\sqrt{2})^{2012}$ is a very small number.} Maybe the final answer is ``9''?\\ \hline
rel[\textbf{generalise}](binomial theorem, eliminate cross terms) & (Section \ref{iatc-examples:gowers}) \textbf{$(\sqrt{2}+\sqrt{3})^{2012} + (\sqrt{2}-\sqrt{3})^{2012}$ is an integer!} \\
\hline
\end{tabular}

\newpage
{\centering
\textbf{Content-Focused Structural Relations}

\par}

\smallskip

\noindent
\noindent\begin{tabular}{|p{.45\textwidth}|p{.45\textwidth}|}
\hline
rel[\textbf{used_in}](pivot_seq, pivot) & 2.~Connecting the dots: At the point where the pivot changes we create a line that passes through the previous \textbf{pivot} and a new \textbf{pivot} -- like a side of a polygon.\\ \hline

rel[\textbf{reform}](H_finite_index_in_G , H_infinite_index_in_G) & (Section \ref{iatc-examples:mathoverflow}, Second comment on question) Yes, the statement is out of focus: $gHg^{-1}$ is intended (\textbf{and ``infinite index case''}). The natural starting point is to ask whether the proof for finite index breaks down. \\ \hline

rel[\textbf{instantiates}]((sqrt(2) + sqrt(3))\textasciicircum2,general form of the problem) & (Section \ref{iatc-examples:gowers}) $(\sqrt{2}+\sqrt{3})^2$ \\ \hline
perf[assert](rel[\textbf{expands}](2 + 2sqrt{2}sqrt{3} + 3, (sqrt(2)+sqrt(3))\textasciicircum2)) & (Section \ref{iatc-examples:gowers}) $2+2\sqrt{2}\sqrt{3}+3$\\ \hline
perf[assert](rel[\textbf{sums}](2 + 2sqrt(2)sqrt(3) + 2 + 2 - 2sqrt(2)sqrt(3)+3,10)) & (Section \ref{iatc-examples:gowers}) $(\sqrt{2}+\sqrt{3})^2 + (\sqrt{2}-\sqrt{3})^2 = 10$ \\ \hline
perf[assert](rel[\textbf{cont_summand}]( (sqrt(2) + sqrt(3))\textasciicircum2012 + (sqrt(2)-sqrt(3))\textasciicircum2012,(sqrt(3)-sqrt(2))\textasciicircum2012) & (Section \ref{iatc-examples:gowers}) And $(\sqrt{3}-\sqrt{2})^{2012}$ is a very small number.
Maybe the final answer is ``9''? \\ \hline
\end{tabular}
\end{document}